%% file: main.tex
\documentclass[10pt,letterpaper]{wam}
\usepackage{lmodern}
\usepackage{array,colortbl}
\definecolor{groupgray}{RGB}{245,246,248}
\definecolor{oursgreen}{RGB}{232,247,236}
\definecolor{baseblue}{RGB}{233,240,255}
\hypersetup{pdftitle={Feed-Forward Gaussian Splatting from Sparse Aerial Views},
  pdfauthor={Dongli Wu, Zhuoxiao Li, Jing OU, Tongyan Hua, Yinrui Ren, Shuai Zhang, Xiaobao Wei, Rongjun Qin, Wufan Zhao},
  hypertexnames=false}
\title{Feed-Forward Gaussian Splatting from Sparse Aerial Views}
\author[1]{Dongli Wu}
\author[1]{Zhuoxiao Li}
\author[1]{Jing OU}
\author[1]{Tongyan Hua}
\author[1]{Yinrui Ren}
\author[1]{Shuai Zhang}
\author[2]{Xiaobao Wei}
\author[3]{Rongjun Qin}
\author[1,\dagger]{Wufan Zhao}

\affiliation[1]{The Hong Kong University of Science and Technology (Guangzhou)}
\affiliation[2]{Peking University}
\affiliation[3]{The Ohio State University}

\contribution[\dagger]{Corresponding author}

\input{sec/0_abstract}
\begin{document}
\maketitle
\begin{figure}[htbp]
\centering
\includegraphics[width=\linewidth]{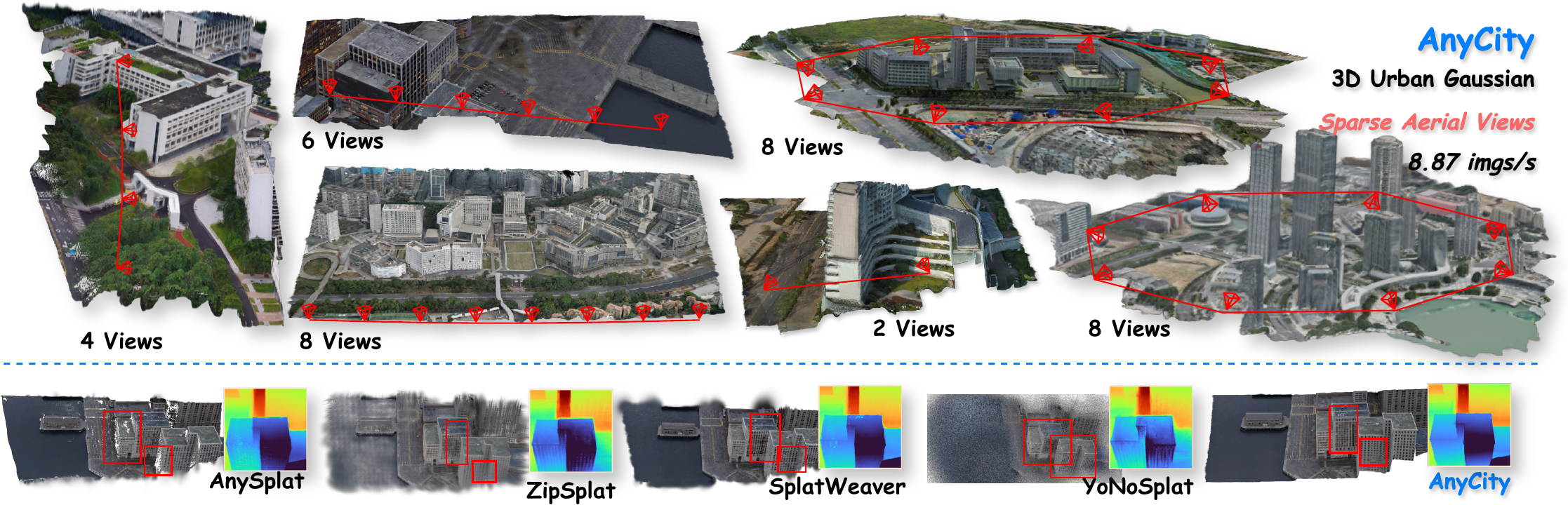}
\caption{
\textbf{From sparse aerial observations to generative urban reconstruction.}
            AnyCity reconstructs 3D Gaussian urban scenes from a few images
            captured along straight or orbiting aerial trajectories
            (\textcolor{red}{red} cameras and paths).
            It produces more complete buildings, clearer facades,
            and cleaner depth boundaries than feed-forward baselines.
        }
        \label{fig:coverfigure}
\end{figure}

\input{sec/1_intro}
\input{sec/2_relatedworks}
\input{sec/3_method}
\input{sec/4_experiment}
\input{sec/5_conclusion}
\FloatBarrier
\input{sec/llm_usage}
\bibliographystyle{corlabbrvnat}
\bibliography{main}
\clearpage
\appendix
\input{appendix}

\end{document}

%% file: sec/0_abstract.tex
\abstract{
Reconstructing cities from a few aerial images can reduce repeat flight passes. Fast reconstruction gives survey crews a preliminary model they can use to decide where to take more images before leaving the site.
Yet aerial data exhibit pose and evidence imbalance, with elevated views favoring roofs and leaving facades and distant structures weakly constrained.
These regions stand to gain most from a video prior, while their geometry is also the hardest to verify from the input views.
To address this, we present \textbf{AnyCity}, a feed-forward generative framework that combines reconstruction and video features to produce a single 3D Gaussian scene in one forward pass.  
\textbf{Aerial Facade Coupling} projects candidate urban surface positions into the input views to guide where and at what scale reconstruction and video features interact. 
We use 3D reference points on the selected surfaces, together with pixel matches across views, to ground geometric updates in the aerial observations.
To isolate the effects of input height and tilt, we introduce \textbf{AnyCity-Aerial}, a benchmark that varies these factors within scenes along parallel and orbiting trajectories and enables comparisons with fixed target views. 
AnyCity outperforms every tested feed-forward baseline fine-tuned on the same aerial training corpus. 
Video-guided refinement yields its largest rendering gains in weakly observed regions, where surface geometry is hardest to recover. 
And constraining the updates with aerial observations improves both novel-view rendering and depth accuracy across the flight altitudes and camera tilts.
}

%% file: sec/1_intro.tex
\section{Introduction}
\label{sec:intro}

\begin{figure}[htbp]
    \centering
    \includegraphics[width=1\linewidth]{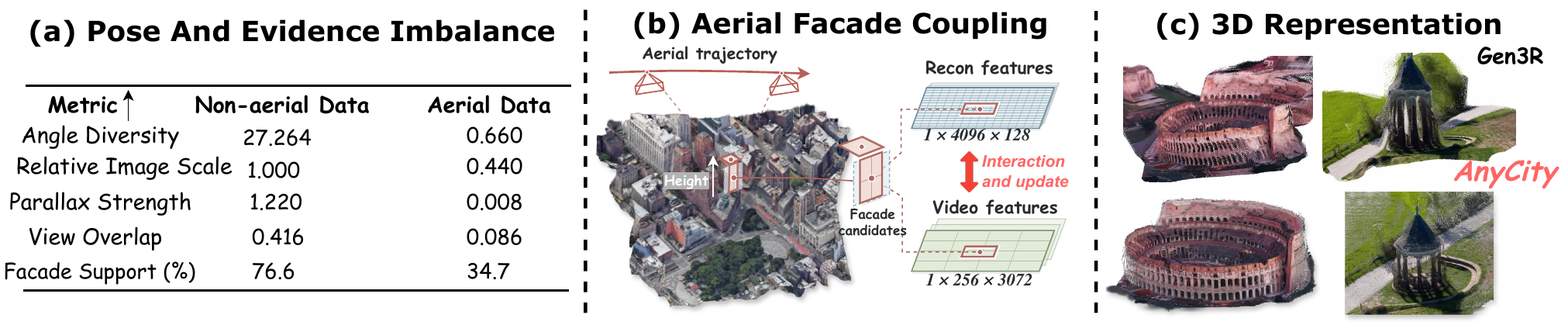}
\caption{
\textbf{(a)} Compared with non-aerial views, sparse aerial observations exhibit pose and evidence imbalance.
\textbf{(b)} Aerial Facade Coupling projects candidate urban surfaces across input views using estimated camera poses to align reconstruction and video features for geometry-constrained interaction and joint updates.
\textbf{(c)} Comparison of 3D representations from Gen3R and AnyCity, with AnyCity achieving more complete and coherent geometry through direct Gaussian decoding.
}
        
    \label{fig:intro}
    
    \end{figure}

Photogrammetry reconstructs 3D scenes from overlapping images for urban mapping and digital twins~\cite{li2023matrixcity}.
At city scale, acquiring dense overlap can require additional flight passes~\cite{haala2011multiray,saadatseresht2026cost}.
Reconstructing from a few views reduces this capture burden.
Fast inference can turn those views into a preliminary model during the survey, giving crews time to identify coverage gaps and capture additional views~\cite{lou2025fly,hanif2025coverage,zhang2024guided}.
This workflow requires a model that can reconstruct quickly from sparse inputs.
City-scale Gaussian methods typically produce detailed reconstructions from dense, calibrated views through time-consuming per-scene optimization~\cite{liu2024citygaussianv2,gao2025citygs,li2026urbangs,yu2026vdgs}.
Feed-forward methods address the input and runtime requirements by predicting a scene directly from a few images in a single pass~\cite{jiang2025anysplat,wang2025vggt,lin2025depth,lu2026grapesplat}.

However, directly applying existing sparse-view feed-forward methods to aerial urban scenes often produces incomplete building structures, blurred facades, and inaccurate depth boundaries, as illustrated in Fig.~\ref{fig:coverfigure}. We attribute these limitations to the \emph{pose and evidence imbalance} inherent in sparse aerial observations (Fig.~\ref{fig:intro} (a)). Unlike close-range views of compact scenes~\cite{kerbl20233d}, aerial views tend to favor rooftops, while facades are often observed only obliquely or from a few viewpoints. At long viewing distances, fine architectural details occupy only a few pixels and may be lost during feature extraction~\cite{xiangli2022bungeenerf}. Limited view overlap further makes it difficult to establish reliable cross-view correspondences. Even when matches are available, restricted viewing-angle diversity and weak parallax leave the 3D positions of building surfaces uncertain. These limitations create two coupled difficulties: recovering missing architectural details and ensuring their geometric consistency with the sparse observations.

Generative priors can reduce those imbalances by drawing on structural patterns learned from large datasets that include large-scale urban scenes~\cite{hua2025sat2city,lee2025skyfall,yu2025orbit}.
Pseudo-view methods use synthesized images as additional supervision~\cite{kong2025generative}, while recent approaches combine video or 3D shape priors with scene reconstruction~\cite{huang2026gen3r,liang2025wonderland,shen2026lyra2,schmid2026genrecon}.
However, generative predictions can inherit errors from the initial geometry or favor plausible shapes that differ from the actual scene (Fig.~\ref{fig:intro} (c)).

To keep the details supplied by the video prior consistent with the actual buildings, we use the aerial cameras' positions and orientations to assess what each view reveals about the same surface.
With \emph{pose and evidence imbalance}, the images can show a surface's texture while leaving its 3D position uncertain.
We use the visible appearance to align details from the video prior with the corresponding part of the urban surfaces.
Where the aerial images provide reliable geometric cues, we use them to constrain the position of the reconstructed scaffold.

To this end, we present \textbf{AnyCity}, a feed-forward generative framework that reconstructs a 3D Gaussian scene from sparse unposed aerial images (Fig.~\ref{fig:coverfigure}). 
The reconstruction backbone builds an initial geometric scaffold and estimates relative heights to guide subsequent feature interactions. The video backbone uses the initial reconstruction features as context to supply complementary appearance cues.

\textbf{Aerial Facade Coupling} guides feature exchange and joint updates between the two streams. 
(a) \textbf{Cross-view feature interaction} determines where to exchange features. We project candidate urban surfaces into the aerial views so
that both streams sample features from corresponding regions of the same
surface across views. (b) \textbf{Joint backbone update} determines how the
exchanged features update both streams. We use cross-view pixel matches and 3D reference points supported by the aerial observations to constrain their shared update coefficients.

After the last video block, the video features provide a final update to the reconstruction state. In the same forward pass, the model outputs a 3D Gaussian scene and estimates a depth map and camera parameters for each input view.

Experiments show that AnyCity outperforms all tested feed-forward baselines fine-tuned on the same aerial data. Video-guided refinement improves rendering most in weakly observed regions. Using cross-view pixel matches and 3D reference points to guide the updates improves rendering and reduces depth error across flight height and camera tilt.

Our contributions are summarized as follows:

\begin{itemize}

    \item \textbf{A feed-forward generative framework for sparse aerial reconstruction} which combines reconstruction and video features to address \emph{pose and evidence imbalance} and limited local detail across aerial urban scenes.

    \item \textbf{Aerial Facade Coupling} guides exchanges between the reconstruction and video streams. It first projects candidate urban surfaces into the input views to align feature interaction, then constrains the joint update using cross-view pixel matches and observation-supported 3D reference points.
    
    \item We introduce \textbf{AnyCity-Aerial}, a benchmark comprising \textbf{6 scenes, 288 sequences, and 18,432 frames} to isolate the effects of flight altitude and camera tilt on aerial reconstruction, obtained along parallel and orbiting aerial trajectories. It provides camera poses and depth maps to systematically evaluate reconstruction across different flight altitudes and camera tilts.
        
\end{itemize}

%% file: sec/2_relatedworks.tex
\section{Related work}

\subsection{Urban scene modeling and synthesis}

Urban reconstruction estimates cameras and geometry from multiple views~\citep{schoenberger2016sfm,schonberger2016pixelwise}, then fits a renderable 3D scene~\citep{mildenhall2021nerf,kerbl20233d,liu2024citygaussianv2,gao2025citygs,li2026urbangs}.
Existing methods address uneven coverage by weighting captured views by visibility or introducing virtual views~\citep{yu2026vdgs,an2026gspotential}.
Generative urban models also restore appearance~\citep{xie2024citydreamer,hua2025sat2city}, for example by adding texture to coarse geometry~\citep{yu2025orbit}.
Sparse aerial images often capture distant structures in only a few pixels and reveal facade details from few angles.
We seek to recover these details with a video prior and use the aerial observations to constrain where they belong.

\subsection{Generalizable 3D reconstruction}

Generalizable models infer cameras and geometry directly from images~\citep{wang2024dust3r,wang2025vggt,lin2025depth}. Feed-forward Gaussian methods extend this approach to renderable scenes from sparse inputs~\citep{ye2024no,zhang2025flare,jiang2025anysplat,lu2026grapesplat}. In aerial imagery, even overlapping views can leave the 3D positions of urban details uncertain. When a prior suggests moving a surface, the available views may provide too little evidence to decide whether the new position is more accurate. AnyCity uses building details visible in multiple aerial images to guide the video prior and keep its refinements consistent with the observed scene.

\subsection{Generative priors for 3D reconstruction}

Generative priors can provide additional cues for 3D reconstruction. Prior work uses synthesized views to supplement supervision or generated images to refine reconstructions~\citep{chen2026gifsplat,wu2024reconfusion,kong2025generative,tang2026vidsplat,liu2026gaussvid,zhang2026dpgs}. Other approaches integrate generation and reconstruction within the model, for example by sharing geometry and appearance representations or by using estimated geometry to guide generation~\citep{chen2026vggt,liang2025wonderland,huang2026gen3r,schmid2026genrecon,shen2026lyra2}. In particular, Gen3R jointly generates aligned geometry and appearance latents through diffusion, while VidSplat and GaussVid use video priors for novel-view synthesis and restoration, respectively. Sparse aerial images often capture buildings from far away and from a limited set of viewpoints. With weak parallax, different 3D shapes can look nearly identical in the input images. Details supplied by a video prior can look realistic and still lie on geometrically inaccurate surfaces. Rather than using video diffusion to generate additional views or the final geometry, AnyCity uses video features to refine an observation-derived reconstruction. Aerial Facade Coupling projects candidate roof and facade surfaces across input views to align feature interaction, and constrains the joint updates using cross-view pixel matches and observation-supported 3D reference points. The updated reconstruction features are then directly decoded into 3D Gaussians in a single forward pass, without iterative denoising or per-scene optimization.

%% file: sec/3_method.tex
\section{AnyCity}
\label{sec:method}

\begin{figure}[htbp]
    \centering
    \includegraphics[width=1\linewidth]{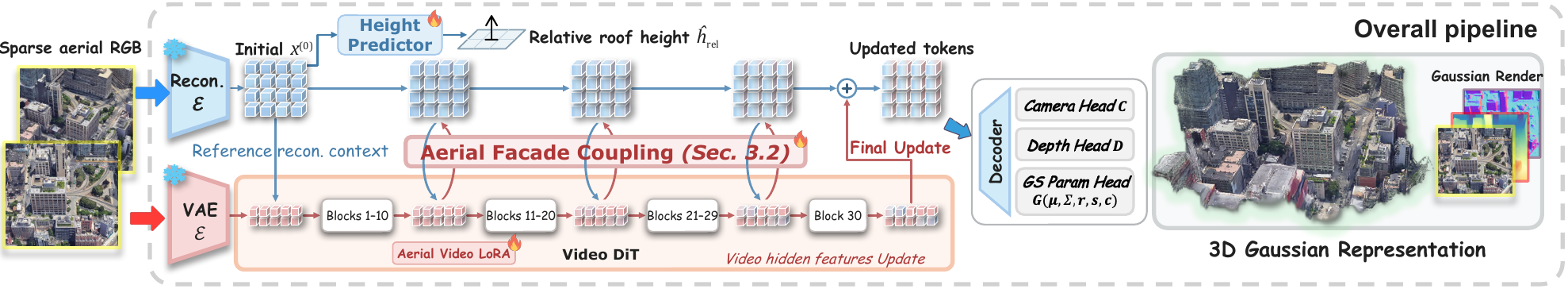}
    \caption{\textbf{Overview of AnyCity.}
    Sparse aerial RGB images enter the reconstruction encoder and the video VAE. The initial reconstruction features provide reference context for the video DiT and feed the height predictor for relative roof height estimation. Aerial Facade Coupling exchanges features and updates both streams after DiT blocks 10, 20, and 29. After block 30, the video stream provides a final update before the reconstruction decoder predicts camera parameters, depth maps, and the 3D Gaussian scene.}
    \label{fig:method_overview}
    
\end{figure}

\paragraph{Overview.}
Given a sparse set of unposed aerial images $\mathcal I_{\mathrm{aerial}}=\{I_i\}_{i=1}^{N}$, AnyCity reconstructs a single 3D Gaussian scene and estimates the camera parameters and depth map for each input view in one forward pass.
Sparse aerial views exhibit \textit{pose and evidence imbalance} (Fig.~\ref{fig:intro}). Even when two views show the same roof clearly, the points we can match across them may provide too little information to determine the building's height. 

The reconstruction backbone's feature grid can also miss fine facade details that occupy only a few pixels in the input images.

AnyCity (Fig.~\ref{fig:method_overview}) uses two streams. The reconstruction stream provides an initial geometric scaffold, and the video stream supplies complementary cues for fine details (Sec.~\ref{sec:obs_latent}). \textbf{Aerial Facade Coupling} determines where the streams interact across aerial views and how they are updated jointly (Sec.~\ref{sec:coupling}). After the last video block, the reconstruction stream reads the updated video features once more before decoding the 3D Gaussian scene. We train AnyCity in two stages, first adapting the reconstruction backbone to aerial geometry, then adapting the video prior and training Aerial Facade Coupling (Sec.~\ref{sec:obs_preserving}).

\subsection{Reconstruction and video backbones}
\label{sec:obs_latent}

\subsubsection{Reconstruction backbone}
We use VGGT-1B~\cite{wang2025vggt} as the reconstruction backbone, retaining its multi-view encoder and its depth and camera heads.
At the start of the forward pass, the backbone jointly encodes the aerial images $\mathcal I_{\mathrm{aerial}}=\{I_i\}_{i=1}^{N}$, producing an initial reconstruction state $x^{(0)}$ and predicting per-view depth maps $D_i^{(0)}$ and camera parameters $\Pi_i^{(0)}$:
\begin{equation}
    \left(x^{(0)},\{(D_i^{(0)},\Pi_i^{(0)})\}_{i=1}^{N}\right)
    =\mathcal E_{\mathrm{rec}}(\mathcal I_{\mathrm{aerial}}).
\end{equation}
Here, $\mathcal E_{\mathrm{rec}}$ denotes the full reconstruction backbone, including its prediction heads.
The state $x^{(0)}$ retains appearance features, intermediate depth features, and camera tokens for later interactions.
To establish a shared 3D reference for cross-view feature exchange, we use the predicted camera parameters to find the viewing ray through pixel $u$ and its predicted depth to locate a point along that ray:
\begin{equation}
    P_i^{(0)}(u)=
    \operatorname{BackProj}\!\left(u,D_i^{(0)}(u);\Pi_i^{(0)}\right).
\end{equation}

The relative-height predictor estimates roof-height offsets from the initial reconstruction features.
For each roof patch, we use its back-projected point as a local 3D reference and the predicted relative height to guide the placement of candidate roof surfaces during cross-view feature exchange (Sec.~\ref{sec:completion_tokens}).
The initial reconstruction features also serve as reference context for the video backbone.

\subsubsection{Video backbone}
The video backbone provides complementary cues for obliquely viewed building facades and distant structures that occupy only a few pixels in the aerial images.
We initialize it from Wan 2.2~\citep{wan2025wan}.
Its image VAE encodes each input image independently, and the resulting latents enter the video diffusion transformer (DiT) at a fixed noise level.
The DiT also receives reference context obtained by projecting the initial reconstruction features.
We denote the DiT hidden state by $h$ and retain the input-view index of each token.
For feature exchange, Aerial Facade Coupling samples from a $32\times32$ reconstruction feature grid and a $16\times16$ video feature grid for each input view.
The video sampling grid is derived from the DiT's $8\times8$ token grid.
To align features from both streams at the same part of a building, we locate the region the surface covers in each image and map that region to the two feature grids.
Sec.~\ref{sec:coupling} explains how we determine these sampling regions and use the sampled features to update both streams jointly.

\subsection{Aerial Facade Coupling}
\label{sec:coupling}

\begin{figure}[htbp]
    \centering
    \includegraphics[width=.82\linewidth]{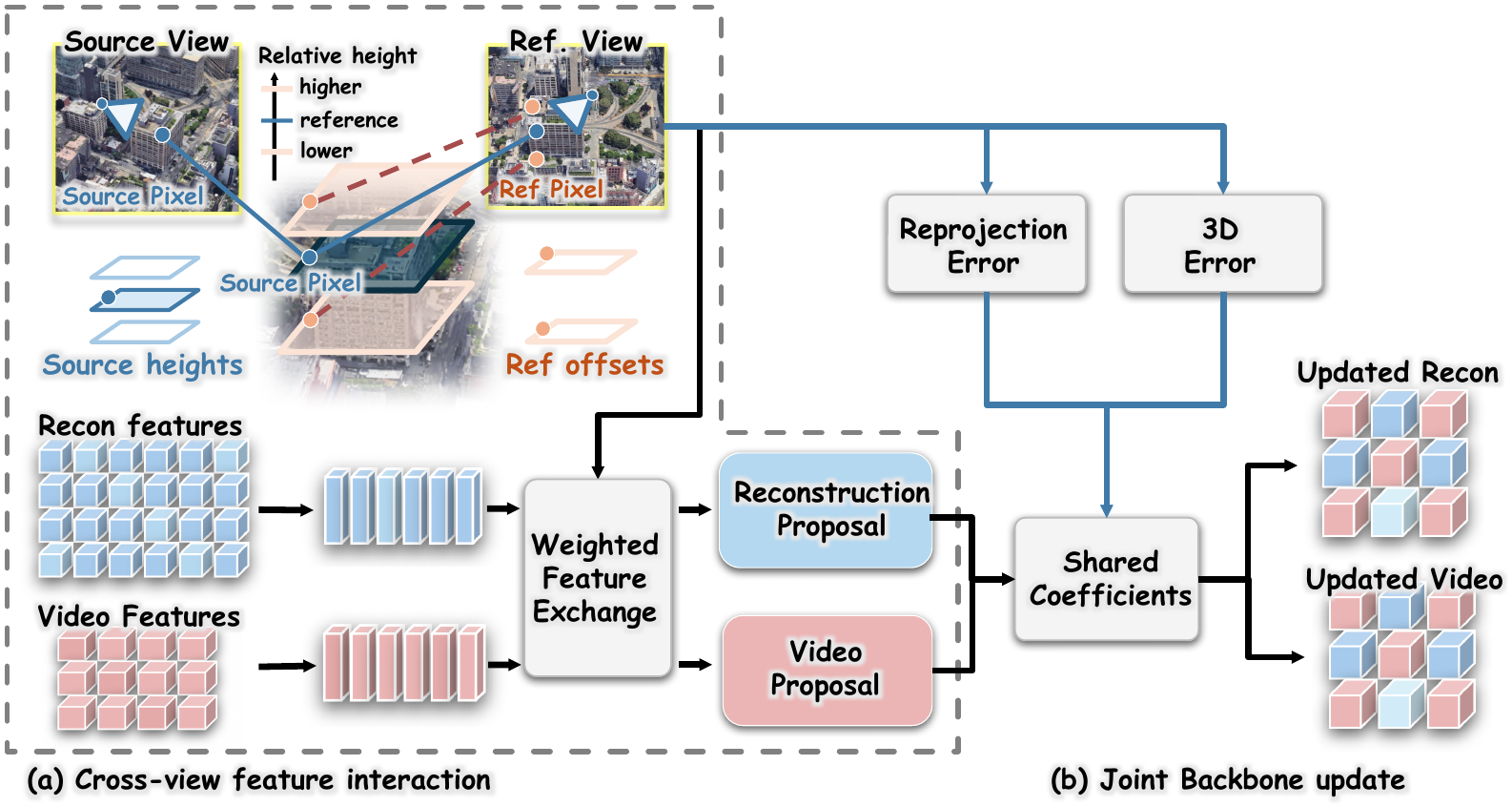}
    \caption{\textbf{Aerial Facade Coupling.} (a) Candidate surfaces map patches across views. Weighted feature samples yield two update proposals. (b) Reprojection and 3D errors guide the shared update coefficients.}
    \label{fig:aerial_facade_coupling}
\end{figure}

An error in the initial depth can misalign a roof patch across views. Aerial Facade Coupling determines where to exchange features by projecting several plausible surface positions and how to update both streams jointly by using aerial observations (Fig.~\ref{fig:aerial_facade_coupling}).

\subsubsection{Cross-view feature interaction}
\label{sec:completion_tokens}

\paragraph{Projecting candidate urban surfaces.}
We start with a pair of matching pixels in two aerial images. Using the back-projected source pixel as an anchor, we place candidate roofs at nearby heights and candidate facades at nearby offsets along the horizontal wall normal. For each candidate, we intersect the rays through the source patch with the candidate plane and project the resulting 3D points into the reference view. The projected patch gives the location and size of the region from which we sample reconstruction and video features.

\paragraph{Combining features and selecting 3D reference points.}
For each candidate, we compare the projected patch center with the observed match in the reference image. This distance penalizes the learned candidate score more strongly when the two views can distinguish the candidates. We combine the feature samples from all valid candidates using weights derived from these scores and the strength of the observations. For example, two roof heights can project to nearly the same patch and both contribute useful roof features. When the views clearly support one surface position over nearby alternatives, we also keep its intersection with the source ray as a 3D reference point. Appendix~\ref{sec:supp_surface_interaction} gives the weights and selection criteria.

\begin{figure}[htbp]
    \centering
    \includegraphics[width=0.9\textwidth]{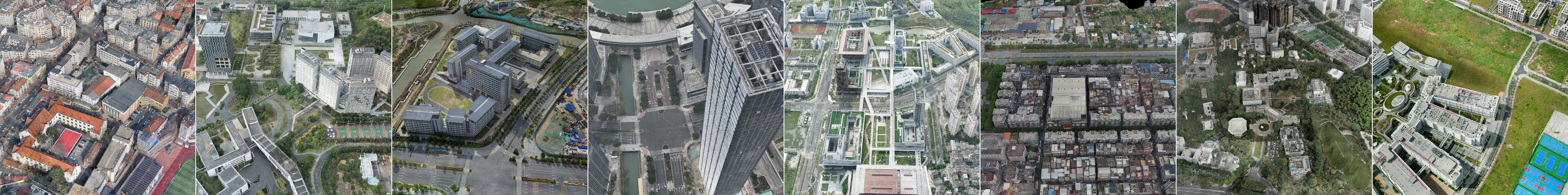}
    \caption{\textbf{AnyCity-Aerial showcase.} Representative aerial views of urban scenes in the benchmark.}
    \label{fig:anycity_aerial_showcase}
    
\end{figure}

\subsubsection{Joint backbone update}
\label{sec:video_prior}

The exchanged features yield a reconstruction proposal $p_r$, a video proposal $p_v$, and paired update directions in the columns of $B_r$ and $B_v$. Each entry of the shared vector $a$ scales one direction in each stream, giving $\Delta x=B_ra$ and $\Delta h=B_va$. This lets geometric measurements on the reconstruction guide both updates.

For a matched pixel pair, we back-project the source pixel and project it into the reference image. The reprojection error is the difference between this projection and the observed match. For a 3D reference point, we project it into an input image and keep that pixel fixed. The 3D error measures how far the back-projection of that pixel is from the reference point. Both errors depend on the depth and cameras predicted from $x$. We linearize them around the current state to estimate the effect of $\Delta x$ on the image and 3D residuals, $r_{\mathrm{img}}(a)$ and $r_{\mathrm{3D}}(a)$.

\paragraph{Solving for shared coefficients.}
We choose $a$ to balance the two proposals against the geometric errors, using $\|z\|_M^2=z^\top Mz$.
\begin{equation}
\label{eq:anycity_objective}
\begin{aligned}
    E(a)={}&\tfrac12\|\Delta x-p_r\|_{M_r}^2
            +\tfrac12\|\Delta h-p_v\|_{M_v}^2\\
           &+\tfrac{\lambda}{2}\|r_{\mathrm{img}}(a)\|_W^2
            +\tfrac{\mu}{2}\|r_{\mathrm{3D}}(a)\|_T^2+\tfrac{\eta}{2}\|a\|_2^2 .
\end{aligned}
\end{equation}
The matrices $M_r$ and $M_v$ account for feature scale. The weights $W$ and $T$ set the contribution of each match and reference point, while $\lambda$ and $\mu$ scale their respective error terms. We penalize large $a$ with $\eta$ and bound $a$ and both updates before applying them (Appendix~\ref{sec:supp_joint_update}).

We update both streams at three points in the DiT. After the last block, the video features provide a final update to the reconstruction before decoding.

\subsection{Progressive training and inference}
\label{sec:obs_preserving}

We train AnyCity in two stages. Stage~I adapts the reconstruction backbone to aerial geometry so that its predictions can anchor the roof and facade candidates used in feature interaction. Stage~II adapts the video prior to aerial images and trains the feature interaction and joint updates. The goal is to use video cues to refine the reconstruction while grounding the updates in matches and 3D reference points supported by the input views. At inference, AnyCity reconstructs a 3D Gaussian scene and predicts per-view camera parameters and depth maps from sparse aerial RGB images in a single forward pass.

\paragraph{Stage I: adapting reconstruction.}
With the image encoder and camera head fixed, we train the depth decoder, appearance feature bridge, Gaussian decoder, and relative-height predictor.
We supervise rendered RGB, depth, normals, and relative roof heights derived from the reference geometry.
With $\mathcal L_{\mathrm{render}}$ denoting the RGB reconstruction loss, the main terms are
\begin{equation}
\label{3}
    \mathcal L_{\mathrm{I}} =
    \mathcal L_{\mathrm{render}} +
    \lambda_D\mathcal L_{\mathrm{depth}} +
    \lambda_N\mathcal L_{\mathrm{normal}} +
    \lambda_H\mathcal L_{\mathrm{height}} .
\end{equation}
The depth term penalizes relative error on valid rendered pixels, the normal term aligns the shortest Gaussian axes with the reference surface orientations, and the height term supervises predicted relative roof heights. Appendix~\ref{sec:supp_stage1} details these losses.

\paragraph{Stage II: adapting video and training the interaction.}
We use LoRA~\citep{hu2021lora} to adapt the video backbone to aerial imagery. We supervise the adaptation with the residual between the clean and noisy image latents.
We then train the video adapters and the interaction modules with the Stage~I reconstruction backbone weights fixed.
The Stage~II objective is
\begin{equation}
\label{stage2}
    \mathcal L_{\mathrm{II}} =
    \mathcal L_{\mathrm{render}} +
    \lambda_P\mathcal L_{\mathrm{pair}} +
    \lambda_A\mathcal L_{\mathrm{coverage}} +
    \mathcal L_{\mathrm{aux}} .
\end{equation}
$\mathcal L_{\mathrm{render}}$ compares the Gaussian RGB renderings with the source images and available held-out training views.
$\mathcal L_{\mathrm{pair}}$ uses rendered depth to reproject matched pixels across views and penalizes their pixel error.
$\mathcal L_{\mathrm{coverage}}$ penalizes low opacity at those pixels.
The weights $\lambda_P$ and $\lambda_A$ control the two terms, respectively.
$\mathcal L_{\mathrm{aux}}$ includes video denoising, feature alignment, an update penalty, and depth, normal, and candidate supervision where geometry labels are available (Appendix~\ref{sec:supp_refinement}).

%% file: sec/4_experiment.tex
\section{Experiments}
\label{sec:experiments}

\subsection{Experimental setup}
\label{sec:experimental_setup}

\noindent\textbf{Training and evaluation.}
Our aerial training corpus combines MatrixCity~\citep{li2023matrixcity}, CityDreamer's GoogleEarth data~\citep{xie2024citydreamer}, and real UAV scenes from UrbanScene3D~\citep{UrbanScene3D}.
Stage I primarily uses MatrixCity for depth and normal supervision. Stage II draws primarily on GoogleEarth and UrbanScene3D for their wider range of aerial viewpoints.
We also fine-tune the feed-forward baselines on the same curated split to control for access to aerial training data.
We evaluate on held-out GoogleEarth and MatrixCity data, UAVScenes~\cite{wang2025uavscenes}, and \textbf{AnyCity-Aerial}.

\noindent\textbf{AnyCity-Aerial benchmark.}
To vary the geometry of the input views within the same scene, we render matched aerial sequences from 3D Gaussian models built from our DJI UAV captures and scenes from GauU-Scene~\cite{xiong2024gauu} and UrbanScene3D~\citep{UrbanScene3D}.
Recorded altitude, gimbal pitch, and GPS tracks guide the parallel and orbiting camera paths.
The benchmark contains six scenes, 288 sequences, and 18,432 frames (Fig.~\ref{fig:anycity_aerial_showcase}).
The matched sequences let us change input height or tilt while keeping the scene, number of inputs, and target cameras fixed.

\definecolor{rankfirst}{RGB}{142,204,168}
\definecolor{ranksecond}{RGB}{185,224,199}
\definecolor{rankthird}{RGB}{224,241,230}

\begin{table}[htbp]
\centering
\setlength\tabcolsep{10pt}
\renewcommand\arraystretch{1}
\caption{\textbf{Quantitative comparison on urban novel view synthesis.}
Dark to light green marks first to third place within each dataset and metric column.}
\label{tab:quantitative}

\raggedright{\small \textbf{(a) Comparison with feed-forward methods}} \\
\centering
\resizebox{\linewidth}{!}{
\begin{tabular}{l *{8}{c!{\hspace{1.2pt}}}c}
\toprule
\multirow{2}{*}{\textbf{Method}} 
& \multicolumn{3}{c}{\textbf{2 Views}} 
& \multicolumn{3}{c}{\textbf{4 Views}} 
& \multicolumn{3}{c}{\textbf{16 Views}} \\
\cmidrule(lr){2-4} \cmidrule(lr){5-7} \cmidrule(lr){8-10}
& PSNR$\uparrow$ & SSIM$\uparrow$ & LPIPS$\downarrow$
& PSNR$\uparrow$ & SSIM$\uparrow$ & LPIPS$\downarrow$
& PSNR$\uparrow$ & SSIM$\uparrow$ & LPIPS$\downarrow$ \\
\midrule

\rowcolor{groupgray}
\multicolumn{10}{l}{\textit{GoogleEarth}} \\
AnySplat & 15.00 & 0.272 & \cellcolor{rankthird}0.376 & 16.01 & 0.302 & 0.353 & 14.86 & 0.206 & 0.508 \\
\rowcolor{black!4}
\quad\textit{AnySplat}$^{\dagger}$ & 15.95 & 0.309 & \cellcolor{ranksecond}0.342 & 17.08 & \cellcolor{rankthird}0.352 & \cellcolor{ranksecond}0.313 & 15.43 & 0.231 & 0.481 \\
SplatWeaver & 15.56 & 0.293 & 0.432 & 16.79 & 0.339 & 0.397 & 15.13 & 0.209 & 0.526 \\
\rowcolor{black!4}
\quad\textit{SplatWeaver}$^{\dagger}$ & \cellcolor{rankthird}16.37 & \cellcolor{ranksecond}0.337 & 0.390 & \cellcolor{rankthird}17.62 & \cellcolor{ranksecond}0.382 & \cellcolor{rankthird}0.352 & 15.82 & \cellcolor{rankthird}0.238 & 0.493 \\
ZipSplat & 16.21 & 0.268 & 0.469 & 16.67 & 0.264 & 0.440 & 14.75 & 0.185 & 0.540 \\
\rowcolor{black!4}
\quad\textit{ZipSplat}$^{\dagger}$ & \cellcolor{ranksecond}17.19 & 0.309 & 0.429 & \cellcolor{ranksecond}17.83 & 0.321 & 0.389 & 15.39 & 0.207 & 0.512 \\
YoNoSplat & 14.27 & 0.276 & 0.461 & 15.92 & 0.303 & 0.433 & 14.13 & 0.196 & 0.503 \\
\rowcolor{black!4}
\quad\textit{YoNoSplat}$^{\dagger}$ & 15.02 & \cellcolor{rankthird}0.312 & 0.414 & 16.43 & 0.333 & 0.398 & 15.61 & 0.201 & 0.479 \\
StructSplat & 15.22 & 0.218 & 0.503 & 15.11 & 0.199 & 0.547 & 12.96 & 0.160 & 0.711 \\
\rowcolor{black!4}
\quad\textit{StructSplat}$^{\dagger}$ & 15.83 & 0.243 & 0.469 & 16.02 & 0.251 & 0.494 & 13.73 & 0.192 & 0.667 \\
Depth Anything 3 & 13.19 & 0.132 & 0.599 & 15.98 & 0.227 & 0.433 & \cellcolor{rankthird}16.06 & 0.223 & \cellcolor{rankthird}0.443 \\
\rowcolor{black!4}
\quad\textit{Depth Anything 3}$^{\dagger}$ & 13.81 & 0.151 & 0.571 & 16.72 & 0.255 & 0.402 & \cellcolor{ranksecond}16.91 & \cellcolor{ranksecond}0.252 & \cellcolor{ranksecond}0.411 \\
\textbf{AnyCity (Ours)} & \cellcolor{rankfirst}\textbf{18.48} & \cellcolor{rankfirst}\textbf{0.361} & \cellcolor{rankfirst}\textbf{0.263} & \cellcolor{rankfirst}\textbf{21.73} & \cellcolor{rankfirst}\textbf{0.496} & \cellcolor{rankfirst}\textbf{0.178} & \cellcolor{rankfirst}\textbf{22.12} & \cellcolor{rankfirst}\textbf{0.489} & \cellcolor{rankfirst}\textbf{0.229} \\
\midrule

\rowcolor{groupgray}
\multicolumn{10}{l}{\textit{UAVScenes}} \\
AnySplat & 20.10 & \cellcolor{rankthird}0.579 & \cellcolor{rankthird}0.236 & 20.62 & \cellcolor{rankthird}0.599 & \cellcolor{rankthird}0.218 & 20.24 & 0.514 & \cellcolor{rankthird}0.263 \\
\rowcolor{black!4}
\quad\textit{AnySplat}$^{\dagger}$ & \cellcolor{ranksecond}20.64 & \cellcolor{ranksecond}0.612 & \cellcolor{ranksecond}0.216 & \cellcolor{rankthird}21.10 & \cellcolor{ranksecond}0.631 & \cellcolor{ranksecond}0.196 & 20.68 & \cellcolor{rankthird}0.535 & \cellcolor{ranksecond}0.245 \\
SplatWeaver & 19.26 & 0.495 & 0.408 & 20.57 & 0.539 & 0.367 & \cellcolor{rankthird}21.05 & 0.530 & 0.333 \\
\rowcolor{black!4}
\quad\textit{SplatWeaver}$^{\dagger}$ & \cellcolor{rankthird}20.15 & 0.538 & 0.368 & \cellcolor{ranksecond}21.35 & 0.585 & 0.328 & \cellcolor{ranksecond}21.63 & \cellcolor{ranksecond}0.556 & 0.309 \\
ZipSplat & 16.64 & 0.382 & 0.519 & 19.03 & 0.444 & 0.438 & 19.41 & 0.440 & 0.405 \\
\rowcolor{black!4}
\quad\textit{ZipSplat}$^{\dagger}$ & 17.71 & 0.423 & 0.478 & 19.86 & 0.491 & 0.391 & 19.96 & 0.465 & 0.381 \\
YoNoSplat & 14.53 & 0.261 & 0.529 & 14.02 & 0.315 & 0.487 & 15.77 & 0.354 & 0.480 \\
\rowcolor{black!4}
\quad\textit{YoNoSplat}$^{\dagger}$ & 15.13 & 0.299 & 0.484 & 15.67 & 0.348 & 0.455 & 16.43 & 0.382 & 0.448 \\
StructSplat & 17.18 & 0.349 & 0.443 & 18.96 & 0.419 & 0.417 & 17.27 & 0.365 & 0.520 \\
\rowcolor{black!4}
\quad\textit{StructSplat}$^{\dagger}$ & 18.15 & 0.399 & 0.405 & 19.31 & 0.449 & 0.390 & 17.78 & 0.387 & 0.487 \\
Depth Anything 3 & 15.47 & 0.240 & 0.519 & 15.83 & 0.246 & 0.501 & 17.57 & 0.342 & 0.415 \\
\rowcolor{black!4}
\quad\textit{Depth Anything 3}$^{\dagger}$ & 16.03 & 0.267 & 0.491 & 16.51 & 0.272 & 0.473 & 18.39 & 0.378 & 0.381 \\
\textbf{AnyCity (Ours)} & \cellcolor{rankfirst}\textbf{22.14} & \cellcolor{rankfirst}\textbf{0.632} & \cellcolor{rankfirst}\textbf{0.213} & \cellcolor{rankfirst}\textbf{24.66} & \cellcolor{rankfirst}\textbf{0.690} & \cellcolor{rankfirst}\textbf{0.186} & \cellcolor{rankfirst}\textbf{24.75} & \cellcolor{rankfirst}\textbf{0.680} & \cellcolor{rankfirst}\textbf{0.201} \\
\bottomrule
\multicolumn{10}{l}{\small $^{\dagger}$ \textit{Fine-tuned on our curated aerial training corpus.}} \\
\end{tabular}
}

\raggedright{\small \textbf{(b) Comparison with per-scene optimization methods}} \\
\setlength\tabcolsep{15pt}
\renewcommand\arraystretch{1}
\centering
\resizebox{\linewidth}{!}{
\begin{tabular}{l *{7}{c!{\hspace{1.2pt}}}c}
\toprule
\multirow{2}{*}{\textbf{Method}} & \multicolumn{4}{c}{\textbf{32 Views}} & \multicolumn{4}{c}{\textbf{64 Views}} \\
\cmidrule(lr){2-5} \cmidrule(lr){6-9}
& PSNR$\uparrow$ & SSIM$\uparrow$ & LPIPS$\downarrow$ & Time$\downarrow$
& PSNR$\uparrow$ & SSIM$\uparrow$ & LPIPS$\downarrow$ & Time$\downarrow$ \\
\midrule

\rowcolor{groupgray}
\multicolumn{9}{l}{\textit{MatrixCity}} \\
CityGaussianV2$^*$ & \cellcolor{rankthird}22.214 & \cellcolor{rankthird}0.674 & \cellcolor{rankthird}0.421 & \cellcolor{rankthird}$\sim$34m & \cellcolor{rankthird}24.536 & \cellcolor{rankthird}0.739 & \cellcolor{rankthird}0.326 & \cellcolor{rankthird}$\sim$49m \\
 
CityGS-X$^*$ & \cellcolor{ranksecond}23.132 & \cellcolor{ranksecond}0.718 & \cellcolor{ranksecond}0.396 & \cellcolor{ranksecond}$\sim$24m & \cellcolor{ranksecond}25.270 & \cellcolor{rankfirst}\textbf{0.783} & \cellcolor{ranksecond}0.307 & \cellcolor{ranksecond}$\sim$36m \\
\midrule
\textbf{AnyCity (Ours)} & \cellcolor{rankfirst}\textbf{25.052} & \cellcolor{rankfirst}\textbf{0.734} & \cellcolor{rankfirst}\textbf{0.357} & \cellcolor{rankfirst}\textbf{3.3s} & \cellcolor{rankfirst}\textbf{25.391} & \cellcolor{ranksecond}0.770 & \cellcolor{rankfirst}\textbf{0.283} & \cellcolor{rankfirst}\textbf{4.4s} \\
\bottomrule
\multicolumn{9}{l}{\small $^*$ \textit{Methods explicitly requiring precise camera poses, depth regularization and initial SfM point clouds.}} \\
\end{tabular}
}
    
\end{table}

\begin{figure}[htbp]
    \centering
    \includegraphics[width=\linewidth]{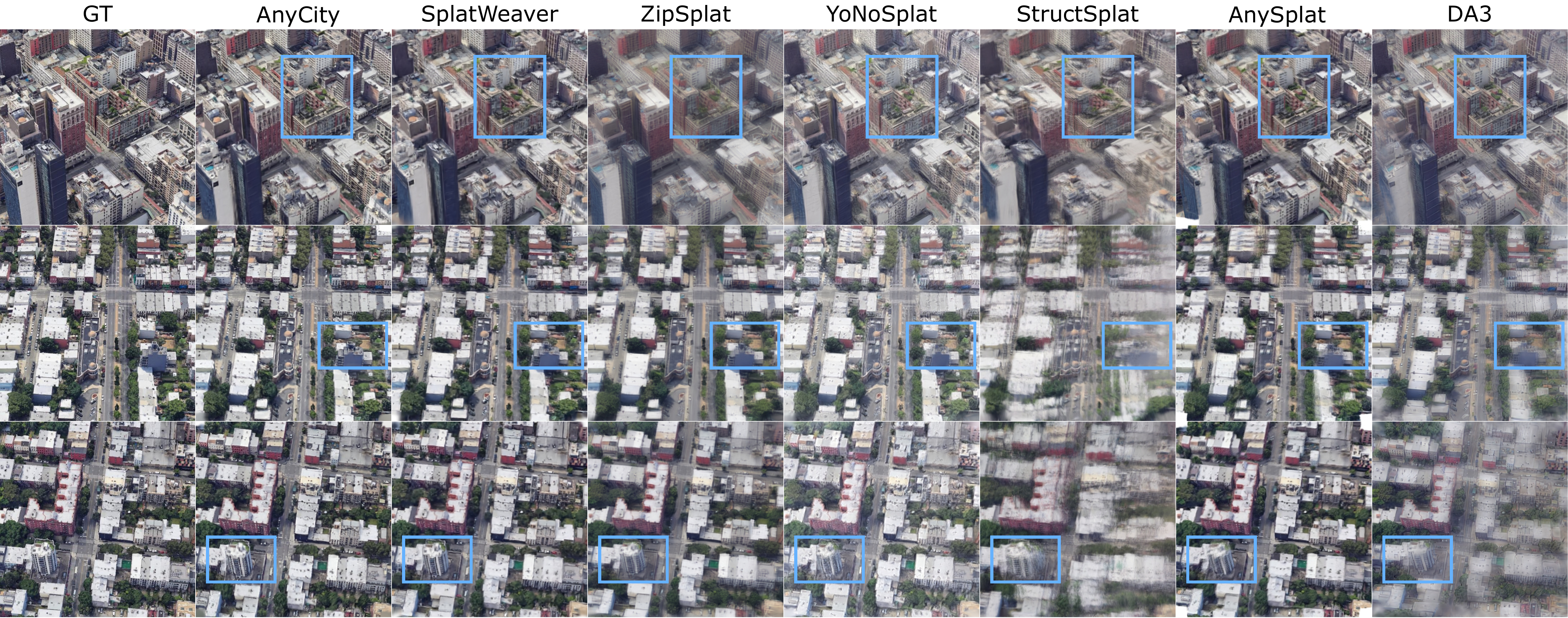}\par\smallskip
    \textbf{(a) Novel-view synthesis with fine-tuned baselines}\par\medskip
    \includegraphics[width=\linewidth]{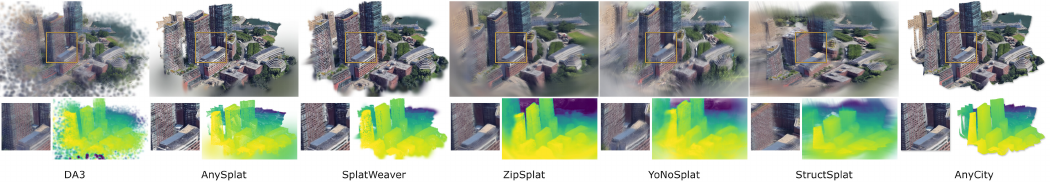}\par\smallskip
    \textbf{(b) Qualitative comparison of 3D reconstructions on GoogleEarth}
    \caption{\textbf{Qualitative comparisons.}
    In (a), each row shows a ground-truth target image and the corresponding renderings. Blue boxes mark regions for comparison.
    (b) Reconstructed 3D scenes and corresponding depth renderings.}
    \label{fig:qualitative_comparison}
    
\end{figure}

\noindent\textbf{Metrics \& Baselines.}
We report PSNR, SSIM~\citep{wang2004image}, and LPIPS~\citep{zhang2018unreasonable} for novel-view rendering, and $\delta_1$~\citep{eigen2015predicting} and AbsRel for depth.
We compare AnyCity with AnySplat~\citep{jiang2025anysplat}, SplatWeaver~\citep{wan2026splatweaver}, ZipSplat~\citep{veicht2026zipsplat}, YoNoSplat~\citep{ye2026yonosplat}, StructSplat~\citep{zhao2026structsplat}, and Depth Anything 3~\citep{lin2025depth} in the sparse feed-forward setting.
We disable ZipSplat's per-scene optimization (TokenBA).
On MatrixCity, we additionally compare with CityGaussianV2~\citep{liu2024citygaussianv2} and CityGS-X~\citep{gao2025citygs} under a denser, per-scene optimization protocol.

\noindent\textbf{Implementation details.}
We use $448\times448$ source crops and initialize the video prior from Wan 2.2~\citep{wan2025wan}.
We use AdamW~\citep{loshchilov2017decoupled} for the training schedule in Sec.~\ref{sec:obs_preserving}.
The three bidirectional updates follow video blocks 10, 20, and 29. The final video-to-reconstruction update follows block 30.
The appendix provides further details on the datasets, training, baseline fine-tuning, and data preparation.

\subsection{Novel view synthesis}

We evaluate how well AnyCity reconstructs urban structures and renders unseen views from sparse aerial images.
On GoogleEarth and UAVScenes, we use the LLFF split (llffhold~=~2) and give all feed-forward methods the same input and target views.
We evaluate each baseline with its original weights and after fine-tuning on the same aerial training corpus.

\noindent\textbf{Sparse feed-forward reconstruction.}
AnyCity achieves the highest PSNR and SSIM and the lowest LPIPS at every tested view count on both datasets (Tab.~\ref{tab:quantitative}(a)).
Fine-tuning improves baseline PSNR by 0.35--1.65~dB, with gains of at most 1~dB in 31 of 36 comparisons.
Fine-tuning helps models adapt to aerial appearance and image scale. For facades seen clearly in only a few images, cross-view matches still provide limited evidence about the surface position.
AnyCity uses a video prior to supply local appearance cues and aerial observations to constrain their placement in 3D.

On GoogleEarth, five of the six baselines lose PSNR when the input count increases from four to sixteen, both before and after fine-tuning. In contrast, AnyCity improves from 21.73 to 22.12 dB, while maintaining its leading performance across all tested view counts. A similar trend is observed on UAVScenes, where AnyCity maintains its PSNR as the input count increases. This contrast highlights the challenges of \emph{pose and evidence imbalance} (Fig.~\ref{fig:intro}). Additional images along the flight path may extend overall coverage while providing limited overlap and parallax to constrain the same surfaces. AnyCity mitigates this issue through geometry-constrained interactions between reconstruction and video features. Tab.~\ref{tab:quantitative}(a) pools results across flight altitudes and camera tilts. Sec.~\ref{sec:anycity_aerial_eval} examines their individual effects at a fixed input view count.

\noindent\textbf{Quality and runtime with more inputs.}
On MatrixCity, we reconstruct scenes from 32 or 64 input views and render the same target views (Tab.~\ref{tab:quantitative}(b)).
The per-scene optimization methods use their required camera poses and initialization, while AnyCity estimates camera parameters from the input RGB images.
In both settings, AnyCity achieves the highest PSNR and lowest LPIPS while substantially reducing reconstruction time.

\noindent\textbf{Qualitative comparison.}
The highlighted regions in Fig.~\ref{fig:qualitative_comparison}(a) show roof outlines and fine building structures.
These details occupy few pixels in the aerial images and appear blurred in several baseline renderings.
The depth maps in Fig.~\ref{fig:qualitative_comparison}(b) show clearer, more coherent building facades in the AnyCity reconstruction.

\begin{table}[htbp]
    \centering
    \caption{\textbf{Effect of input flight geometry on AnyCity-Aerial.}
    Each setting uses four inputs.
    Each cell reports PSNR$\uparrow$ / Obs.~AbsRel$\downarrow$ / Weak~AbsRel$\downarrow$.
    $\dagger$ marks fine-tuned baselines.}
    \label{tab:anycity_aerial_eval}
    \footnotesize
    \setlength{\tabcolsep}{4pt}
    \noindent\makebox[\linewidth][l]{\textbf{(a) Input altitude}}\par\smallskip
    \begin{tabular*}{\linewidth}{@{\extracolsep{\fill}}lcc@{}}
        \toprule
        Method & 150~m & 250~m \\
        \midrule
        \textit{SplatWeaver}$^{\dagger}$ & 16.63 / 0.145 / 0.341 & 16.09 / 0.165 / 0.402 \\
        \textit{AnySplat}$^{\dagger}$ & 16.27 / 0.158 / 0.365 & 15.70 / 0.179 / 0.426 \\
        AnyCity ($\lambda=\mu=0$) & 19.64 / 0.113 / 0.239 & 19.04 / 0.130 / 0.298 \\
        \textbf{Full AnyCity} & \textbf{20.08 / 0.104 / 0.205} & \textbf{19.57 / 0.117 / 0.257} \\
        \bottomrule
    \end{tabular*}
    \par\medskip
    \noindent\makebox[\linewidth][l]{\textbf{(b) Input tilt}}\par\smallskip
    \begin{tabular*}{\linewidth}{@{\extracolsep{\fill}}lcc@{}}
        \toprule
        Method & $0^\circ$ & $45^\circ$ \\
        \midrule
        \textit{SplatWeaver}$^{\dagger}$ & 16.53 / 0.172 / 0.421 & 16.03 / 0.148 / 0.356 \\
        \textit{AnySplat}$^{\dagger}$ & 16.18 / 0.186 / 0.446 & 15.62 / 0.162 / 0.379 \\
        AnyCity ($\lambda=\mu=0$) & 19.55 / 0.133 / 0.311 & 18.98 / 0.115 / 0.252 \\
        \textbf{Full AnyCity} & \textbf{19.97 / 0.120 / 0.267} & \textbf{19.46 / 0.106 / 0.217} \\
        \bottomrule
    \end{tabular*}
    
\end{table}

\begin{table}[htbp]
    \centering
    \caption{\textbf{Ablation study.}
    Obs. and Weak denote observation-supported and weakly observed regions.}
    \label{tab:ablation}
    \footnotesize
    \renewcommand\arraystretch{1.05}
    \noindent\makebox[\linewidth][l]{\textbf{(a) Video-guided refinement by region}}\par\smallskip
    \setlength\tabcolsep{3pt}
    \begin{tabular*}{\linewidth}{@{\extracolsep{\fill}}lccc@{}}
        \toprule
        \textbf{Model configuration}
        & \shortstack{Obs.PSNR$\uparrow$}
        & \shortstack{Weak PSNR$\uparrow$}
        & \shortstack{Weak LPIPS$\downarrow$} \\
        \midrule
        Stage I (Recon. Only) & 21.48 & 17.83 & 0.304 \\
        \textbf{Full AnyCity} & \textbf{23.62} & \textbf{20.26} & \textbf{0.266} \\
        \bottomrule
    \end{tabular*}
    \par\medskip
    \noindent\makebox[\linewidth][l]{\textbf{(b) Training and coupling ablations}($\delta_1$ in \%) }\par\smallskip
    \setlength\tabcolsep{1.5pt}
    \begin{tabular*}{\linewidth}{@{\extracolsep{\fill}}p{0.42\linewidth}ccccc@{}}
        \toprule
        \textbf{Model configuration}
        & \textbf{PSNR}$\uparrow$
        & \textbf{SSIM}$\uparrow$
        & \textbf{LPIPS}$\downarrow$
        & $\delta_1\ \uparrow$
        & \textbf{AbsRel}$\downarrow$ \\
        \midrule
        \multicolumn{6}{l}{\textit{Training}} \\
        w/o aerial video LoRA & 20.38 & 0.457 & 0.218 & 92.7 & 0.127 \\
        w/o depth loss & 19.92 & 0.441 & 0.239 & 80.4 & 0.168 \\
        \midrule
        \multicolumn{6}{l}{\textit{Aerial Facade Coupling}} \\
        only Cross-Attention & 19.21& 	0.432	&0.235	&91.5	&0.136 \\
        Single-position exchange & 20.79 & 0.471 & 0.207 & 92.3 & 0.131 \\
        One-way updates & 21.06 & 0.483 & 0.192 & 93.1 & 0.124 \\
        w/o reprojection ($\lambda=0$) & 20.94 & 0.479 & 0.199 & 90.5 & 0.140 \\
        w/o 3D point term ($\mu=0$) & 20.87 & 0.475 & 0.204 & 89.9 & 0.143 \\
        \midrule
        \textbf{Full AnyCity} & \textbf{21.73} & \textbf{0.496} & \textbf{0.178} & \textbf{93.8} & \textbf{0.120} \\
        \bottomrule
    \end{tabular*}
    
\end{table}

\subsection{Effect of input flight geometry}
\label{sec:anycity_aerial_eval}

Flight altitude and camera pitch affect surface visibility and parallax, which in turn influence depth estimation.
To isolate these factors, we use paired aerial sequences of the same scenes in AnyCity-Aerial and compare altitudes of 150 and 250~m and viewing angles of $0^\circ$ (nadir) and $45^\circ$ (oblique) in separate experiments.
Both experiments use four input images per setting.
We compare with two of the stronger fine-tuned baselines.
To assess the role of geometric constraints, we also evaluate an inference variant using the same AnyCity weights.
In the full model, Aerial Facade Coupling uses the reprojection error at cross-view pixel matches and the displacement of reconstructed points from 3D reference points to solve for joint updates to both streams.
The variant in Tab.~\ref{tab:anycity_aerial_eval} sets $\lambda=\mu=0$ in Eq.~(\ref{eq:anycity_objective}) to omit both error terms from the joint update.

We align the predicted input camera poses to the benchmark poses and render at the same target views.
For points with valid reference depth, we classify a point as observation-supported if it is visible from at least two input views with sufficient parallax in both settings of a comparison.
The remaining valid points are classified as weakly observed.
All methods use the same region masks within each comparison.
We report full-image PSNR and depth AbsRel for each region.
In Tab.~\ref{tab:anycity_aerial_eval}, higher altitude reduces PSNR and increases depth error for every method.
The nadir inputs yield higher PSNR, and the oblique inputs yield lower AbsRel in both regions.
Nadir views mainly show roofs. Oblique views expose more facades, which can provide additional depth cues and make fine details harder to render.
Full AnyCity achieves the highest PSNR and the lowest AbsRel in across all settings.
Adding the geometric constraints reduces depth error more in weakly observed regions, showing that the constraints help recover more accurate surface depths where aerial observations provide limited evidence.

\subsection{Ablation study}
\label{sec:ablation}

We evaluate all variants on the same held-out GoogleEarth subset. The reference depth maps used for evaluation are from MoGe2-Aerial~\citep{song2026aerialmetric} using a checkpoint specialized for aerial imagery. 

\noindent\textbf{Where and how does video-guided refinement help?}
We compare full AnyCity with a Stage I reconstruction-only baseline and a variant using standard cross-attention instead of Aerial Facade Coupling.
As shown in Tab.~\ref{tab:ablation}, video-guided refinement brings larger gains in weakly observed regions, while cross-attention alone performs worse than our geometry-constrained coupling.
These results highlight the importance of both generative priors and observation-guided feature interaction.

\noindent\textbf{How do aerial video adaptation and depth loss help?}
Removing aerial video LoRA reduces rendering quality. Removing the depth loss yields the highest depth AbsRel among the variants in Tab.~\ref{tab:ablation}(b).

\noindent\textbf{Do multiple candidate reads improve feature exchange?}
We restrict feature exchange to the initial surface position and retain candidate scoring and the 3D reference points used by the solver. Depth AbsRel increases from 0.120 to 0.131 (Tab.~\ref{tab:ablation}(b)), supporting the use of multiple plausible positions for the feature read.

\noindent\textbf{What do the two-way and geometry-constrained updates contribute?}
With one-way updates, the reconstruction stream reads video features and updates its own state. The video state stays fixed at the coupling layers. The lower PSNR and higher depth error suggest that passing the updated video state to subsequent interactions improves reconstruction.
We separately remove the reprojection term (\(\lambda=0\)) and the 3D point term (\(\mu=0\)) from Eq. (3). The reprojection term measures the discrepancy between a pixel reprojected into another view and its observed match. The 3D point term penalizes deviations of reconstructed points from 3D reference points supported by the input views. Removing either term increases depth error.

%% file: sec/5_conclusion.tex
\section{Conclusion}

We presented AnyCity, a feed-forward framework for reconstructing 3D Gaussian scenes from sparse, unposed aerial images, and introduced AnyCity-Aerial to isolate the effects of flight altitude and camera tilt. AnyCity uses aerial observations to guide where the reconstruction and video streams exchange features and how the two streams are updated jointly. Video-guided refinement improves appearance most in weakly observed regions to address the pose and evidence imbalance in sparse aerial reconstruction.

%% file: sec/llm_usage.tex
\section*{LLM Usage Statement}
We used OpenAI GPT to discuss research ideas and relevant literature, polish the manuscript's language, organize its structure and layout, search for relevant literature, and flag apparent errors in our claims.

%% file: appendix.tex
\setlength{\parskip}{4pt plus 1pt minus 1pt}
\titlespacing*{\subsection}{0pt}{9pt plus 1pt minus 1pt}{3pt}
\setcounter{figure}{0}
\setcounter{table}{0}
\setcounter{equation}{0}
\renewcommand{\thefigure}{S\arabic{figure}}
\renewcommand{\thetable}{S\arabic{table}}
\renewcommand{\theequation}{S\arabic{equation}}

\section*{Supplementary Material}
\phantomsection
\addcontentsline{toc}{section}{Supplementary Material}

% Appendix-only contents: independent of the main-paper section list.
\subsection*{Contents}
\begingroup
\newcommand{\suppcontentsentry}[2]{%
  \noindent\hyperref[#1]{\ref*{#1}\quad #2}%
  \nobreak\dotfill\hyperref[#1]{\pageref*{#1}}\par\smallskip}
\suppcontentsentry{sec:supp_coupling}{Method details}
\suppcontentsentry{sec:supp_experimental_settings}{Experimental details}
\suppcontentsentry{sec:supp_questions}{Discussion}
\suppcontentsentry{sec:supp_dataset}{Dataset presentation}
\suppcontentsentry{sec:supp_extra}{Additional experiments}
\suppcontentsentry{sec:supp_discussion}{Limitations and future work}
\suppcontentsentry{sec:supp_component_lineage}{Relation to Prior Work}
\endgroup

\section{Method details}
\label{sec:supp_coupling}

\begin{figure}[htbp]
  \centering
  \includegraphics[width=\linewidth,height=.65\textheight,keepaspectratio]{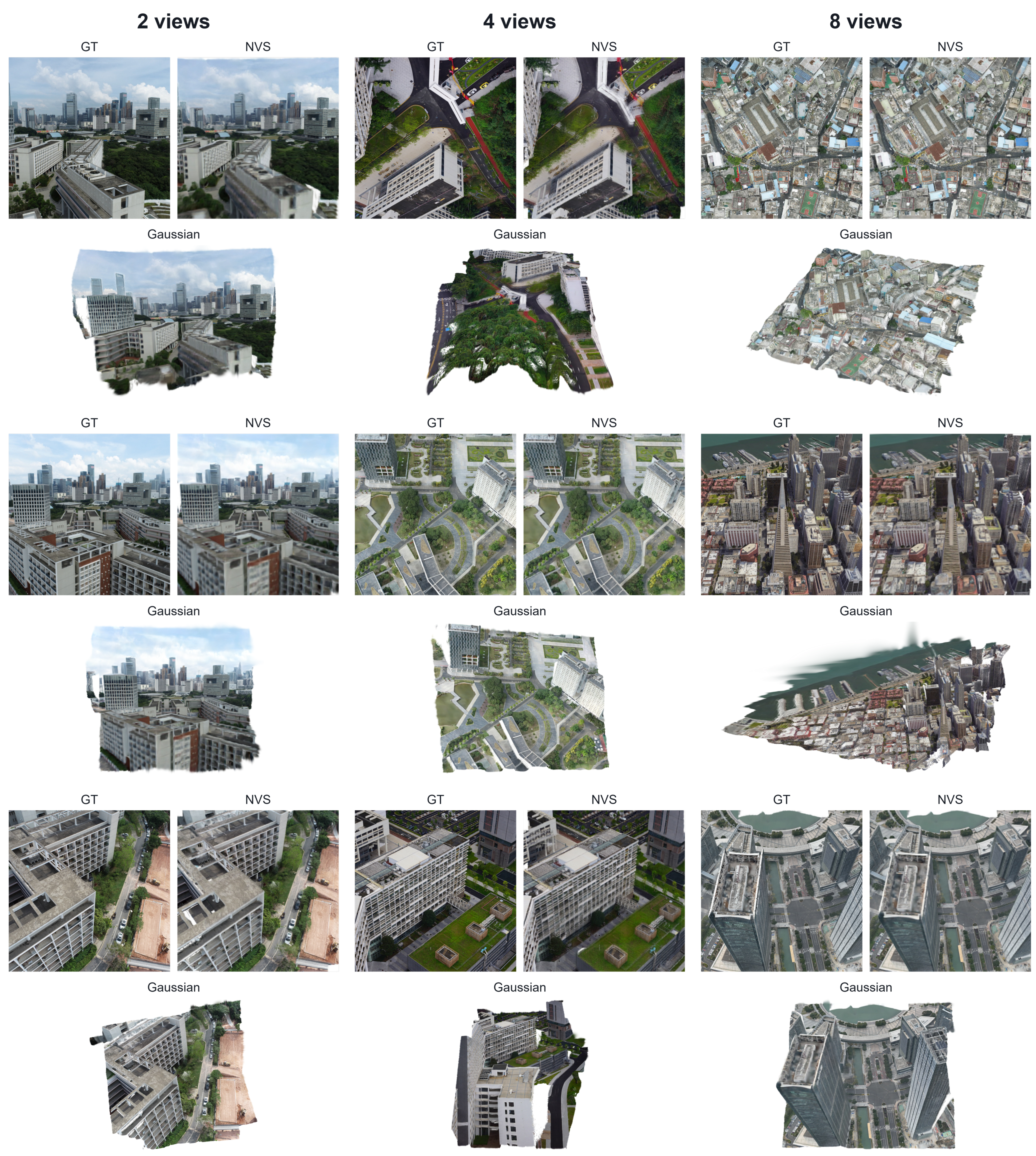}
  \caption{\textbf{AnyCity reconstructions from 2, 4, and 8 input views.}
  Each example shows a ground-truth target image (GT), the corresponding
  novel-view synthesis (NVS), and the reconstructed 3D Gaussian scene.}
  \label{fig:supp_input_view_examples}
\end{figure}

To reconstruct urban geometry and render novel views from sparse aerial inputs (Fig.~\ref{fig:supp_input_view_examples}), AnyCity combines a reconstruction stream that provides the initial geometry with a video stream that supplies fine appearance cues (Sec.~\ref{sec:obs_latent}). Aerial Facade Coupling (AFC) connects the two streams and uses the input views to guide how they exchange features and update the reconstruction (Sec.~\ref{sec:coupling}). Below, we detail how AFC projects roof and facade candidates to sample features across views and uses the available geometric evidence to constrain updates to both streams.

\subsection{Cross-view feature interaction}
\label{sec:supp_surface_interaction}

AFC starts with a pixel match $(u_i,u_j)$ between source view $i$ and reference view $j$. It then projects roof and facade candidates into view $j$ and uses the match to weight the sampled features. When the views clearly favor one candidate, AFC retains its 3D intersection as a reference point (Sec.~\ref{sec:completion_tokens} and Fig.~\ref{fig:aerial_facade_coupling}(a)).

\paragraph{1. Pixel matching and view pairs.}
We compute the matches once per forward pass, before the first AFC update. For each ordered pair of distinct input views, we use the initial depth and cameras to project source pixels with valid depth into the reference view. When a projection is inside the image and has positive depth, we compare the source reconstruction feature with features in a $3\times3$ window centered on the projection using cosine similarity. We keep matches with a forward--backward error of at most one feature cell.

Let $v$ denote a valid location in the window, measured in reference-crop pixels. We turn the similarity scores into weights $q_v$ using a softmax with temperature $0.1$. The matched location is the weighted mean $u_j=\sum_v q_v v$, with covariance
\[
 \Sigma_j = \sum_v q_v(v-u_j)(v-u_j)^\top + (1\,\mathrm{px})^2 I_2.
\]
We skip empty windows and subtract the highest score before computing the softmax. We compute covariance-weighted distances in FP32 using a Cholesky solve. All three AFC updates reuse the same matches and covariances.

A correct match inside the window gives AFC a signal to correct small projection errors. If the true correspondence lies outside the window, the local search misses it. Repeated textures can still produce a false match that passes the forward--backward check. At later AFC updates, we use the updated depth and cameras to project the surface candidates again while reusing the original matches.

\paragraph{2. Project candidate roofs and facades (urban surfaces).}
We back-project the source pixel $u_i$ with the predicted depth to obtain a local 3D anchor. We keep the observed match $u_j$ fixed when comparing candidates.
We compute each point-map normal by taking the cross product of horizontal and vertical central differences, using valid neighboring 3D points.

We find the dominant axis of the initial point-map normals in a common camera frame. We orient that axis toward the cameras to define the scene's up direction. The median valid depth in the initial maps defines the scene scale $L$.

The current surface normal guides the choice of plane families. For roughly horizontal patches, we place roof planes around the predicted height relative to the anchor. When the normal has a clear horizontal component, we normalize it to define the facade normal and shift the facade planes along it from the anchor. In both families, the spacing scales with $L$.

For a plane with normal $n_k$ through point $c_k$, the ray from camera center $o_i$ along unit direction $v_i(u_i)$ hits the plane at $X_k$ and projects to $u_{j,k}$ in view $j$:
\begin{equation}
\begin{split}
 X_k &= o_i+
 \frac{n_k^\top(c_k-o_i)}{n_k^\top v_i(u_i)}v_i(u_i),\\
 u_{j,k} &= \pi_j(X_k;x).
\end{split}
\label{eq:anycity_transfer}
\end{equation}
Projecting the whole patch gives the sampling region in view $j$.
We discard near-parallel rays, intersections behind either camera, and regions outside the valid image.
We map the region to the $32\times32$ reconstruction grid and the $16\times16$ video sampling grid using the same normalized image coordinates.
The video sampling grid is derived from the $8\times8$ DiT token grid.

Table~\ref{tab:supp_afc_configuration} summarizes how AFC builds and scores the candidates.
\begin{table}[htbp]
  \centering
  \caption{\textbf{AFC candidate construction and scoring.}}
  \label{tab:supp_afc_configuration}
  \small
  \setlength{\tabcolsep}{3pt}
  \begin{tabular}{@{}p{0.37\linewidth}>{\raggedright\arraybackslash}p{0.58\linewidth}@{}}
    \toprule
    \textbf{Step} & \textbf{Cue} \\
    \midrule
    Roof planes & Local surface orientation and predicted roof height \\
    Facade planes & Horizontal component of the local surface normal \\
    Plane spacing & Offsets scaled by median scene depth \\
    Mode grouping & Intersections close along the source ray, relative to scene scale \\
    Candidate scoring & Features at the source patch and the projected region \\
    \bottomrule
  \end{tabular}
\end{table}

\paragraph{3. Weight the feature samples.}
The offset from the projected center $u_{j,k}$ to the observed match $u_j$ tells us how well a candidate fits the match.
Its influence on the feature weight depends on how well the views can distinguish the candidates.
For example, if two roof heights project to nearly the same patch, both can provide useful features even though the match gives little evidence about which height is correct.

To compute this observation support for a family $f$ (roof or facade), let $\mathcal K_f$ be its valid candidates.
For a candidate pair $(k,\ell)$, let $\Delta_{k\ell}$ be the distance between their projected centers in cells of the $16\times16$ grid.
Let $\omega_{k\ell}$ be the square root of the smallest absolute normal--ray dot product across those candidates and both views.
Observation support increases with projected separation and more direct viewing angles:
\begin{equation}
 \rho_f=\max_{\substack{k,\ell\in\mathcal K_f\\k<\ell}}
 \frac{\Delta_{k\ell}}{1+\Delta_{k\ell}}\omega_{k\ell}.
\label{eq:supp_support}
\end{equation}
With fewer than two valid candidates, $\rho_f=0$. If the camera centers coincide, points on the same source ray project to the same location in the reference view.

For each candidate, we average the feature samples inside its projected footprint in each stream. We concatenate these two averages and the source reconstruction feature and apply a two-layer MLP with 64 hidden units and GELU to produce the scalar score $b_k$.

We subtract a penalty from the learned score $b_k$ when the projected center is far from the match. The family support $\rho_{f(k)}$ controls how strongly the match error $d_k$ affects that score.
The covariance $\Sigma_j$ describes the match uncertainty in view $j$. We measure both $u_{j,k}$ and $u_j$ in reference-crop pixels and use $\|z\|_M^2=z^\top Mz$. The penalty gives more weight to errors along directions where the match is more precise.
\begin{equation}
\begin{split}
 d_k&=\log\!\left(1+\|u_{j,k}-u_j\|_{\Sigma_j^{-1}}^2\right),\\
 p_k&=\operatorname{softmax}_k\!\left(b_k-\tfrac12\rho_{f(k)}d_k\right).
\end{split}
\label{eq:anycity_candidate_score}
\end{equation}
The softmax gives a probability $p_k$ for each valid candidate. For the feature exchange, we sum the samples as $\sum_k w_k F_k$, with $w_k=p_k(1+\rho_{f(k)})/2$. This weighted sum preserves the reduction in feature strength when observation support is weak.

\paragraph{4. Select a 3D reference point.}
We sort the candidate intersections $X_k$ along the source ray and group nearby points into \emph{modes}. Each group's maximum span scales with $L$. The probability mass of mode $\mathcal{M}$ is $m(\mathcal{M})=\sum_{k\in\mathcal{M}}p_k$. Its representative is the candidate with the largest $p_k$.

\noindent\textbf{Can the views tell the positions apart?}
Take the two most probable modes. Their representatives' projections $\hat u_1$ and $\hat u_2$ in view $j$ must be farther apart than the pixel match uncertainty. This means $\|\hat u_1-\hat u_2\|_{\Sigma_j^{-1}}>1$.

\noindent\textbf{Is the lead strong enough?}
Let $m_1$ and $m_2$ be the leading and runner-up mode masses. We compute the support $\rho_{12}$ from the pairwise term in Eq.~\ref{eq:supp_support}, using the two mode representatives and their surface normals. We use this support to scale the gap between the mode masses.
This weighted gap must outweigh the total mass of all other modes:
\begin{equation}
 \rho_{12}(m_1-m_2)>1-m_1.
\label{eq:supp_adaptive_reference}
\end{equation}
Weaker observation support calls for a larger lead.

If both checks pass, $X^\star$ is where the source ray meets the winning candidate plane. With fewer than two modes or a failed check, we skip the reference point and keep any valid feature samples.
Section~\ref{sec:supp_joint_update} uses retained points to form the 3D residual.

\subsection{Joint backbone update}
\label{sec:supp_joint_update}

Section~\ref{sec:supp_surface_interaction} aligns features across views and selects 3D reference points. In the joint update, each stream uses the aligned features to propose a change. AFC balances these proposals against reprojection and 3D errors to solve for shared update coefficients. It then limits the size of both updates before applying them (Sec.~\ref{sec:video_prior} and Fig.~\ref{fig:aerial_facade_coupling}(b)).

\paragraph{1. Check the geometry.}
\textbf{Pixel matches.} We lift a source pixel $u_i$ into 3D using the predicted depth and camera, then project it into view $j$. The difference between where it lands and the observed match $u_j$ is the image error:
\begin{equation}
 e_{i\to j}(x)=\pi_j(P_i(u_i;x);x)-u_j.
\end{equation}
Each accepted ordered match contributes one residual. The ordered pairs include both view directions.

\noindent\textbf{3D reference points.} At the current state $\bar x$, we project $X^\star$ into view $i$ and fix that pixel as $\bar u_i=\pi_i(X^\star;\bar x)$. The error is the offset from $X^\star$ to the reconstructed point at $\bar u_i$:
\begin{equation}
 q_i(x)=P_i(\bar u_i;x)-X^\star.
\label{eq:anycity_update_residuals}
\end{equation}
During linearization, the depth and camera predictions vary with $x$. The matches and covariances, selected 3D points and their projected pixels, and residual weights stay fixed for that solve.

\paragraph{2. Update parameterization.}
At each source location, we average the weighted features from the other stream over reference views with valid samples. Locations without valid samples receive zero context. We use bilinear interpolation to map this context to the receiving grid and a per-view mean for camera tokens. For each appearance, depth, camera, or video group, an MLP combines the context with the current tokens to produce a proposal and update directions. Stacking these outputs gives the proposals $p_r,p_v$ and the input-dependent direction pairs $(d_{r,k},d_{v,k})$.

The feature metric $M_s$ averages the squared update magnitudes across the $G_s$ groups of stream $s\in\{r,v\}$ after accounting for each group's feature scale. For group $g$ with $d_g$ entries, let $\sigma_g$ be the root mean square (RMS) of its current features, floored at $10^{-3}$. The corresponding block of $M_s$ is $I_{d_g}/(G_s d_g\sigma_g^2)$. We normalize both directions in pair $k$ by the same factor
\[
 \max\!\left(10^{-6},\sqrt{
 d_{r,k}^{\top}M_r d_{r,k}+d_{v,k}^{\top}M_v d_{v,k}}\right)
\]
to obtain column $k$ of $B_r$ and $B_v$. The shared solve has $K=4$ coefficients, one for each direction pair.

Let $e,q$ be the stacked residuals at $\bar x$ and $J_e,J_q$ their Jacobians at that state. The linearized residuals are $e+Aa$ and $q+Ca$, where $A=J_eB_r$ and $C=J_qB_r$.

\paragraph{Directional products.}
We use one forward-mode Jacobian--vector product (JVP) per column of $B_r$ to compute $A$ and $C$. We batch the JVPs from the existing state through the depth and camera readouts and the projection operations. The solve needs only these directional derivatives.

\paragraph{Residual weights.}
For the $N_e$ directed pixel residuals, each block of $W$ is $\Sigma_j^{-1}/N_e$. For the $N_q$ 3D residuals, each block of $T$ is $cI_3/[N_q(0.02L)^2]$. The factor $c=\rho_{12}(m_1-m_2)$ weights each point by the support-weighted gap of its selected mode. For full AFC, we use $\lambda=\mu=0.1$ and $\eta=10^{-3}$.

Equation~\ref{eq:anycity_objective} balances the feature proposals against the geometric errors and penalizes large coefficients. With the linearized residuals, we solve $H\hat a=b$, where
\begin{equation}
\begin{split}
 H={}&B_r^\top M_rB_r+B_v^\top M_vB_v\\
 &+\lambda A^\top WA+\mu C^\top TC+\eta I,\\
 b={}&B_r^\top M_rp_r+B_v^\top M_vp_v\\
 &-\lambda A^\top We-\mu C^\top Tq.
\end{split}
\label{eq:supp_normal_system}
\end{equation}
Positive-semidefinite metrics, nonnegative geometric weights, and $\eta>0$ make $H$ positive definite. If there are no retained points, we omit the 3D terms. If there are no matches, we omit the image terms.
We solve the system by Cholesky factorization in FP32. During training, gradients pass through the solve to the proposals, bases, and residuals. The JVP outputs $A,C$, observations, residual weights, and RMS scales are detached, so backpropagation avoids second-order derivatives of the readouts.

\paragraph{3. Limit and apply the updates.}
We apply one scale $\gamma$ to satisfy the bound $r_a=0.25$ on the coefficient norm and all group bounds $r_g=0.1\,\sigma_g\sqrt{d_g}$ on the update norms. In the expression below, $B=[B_r^\top,B_v^\top]^\top$ stacks the two bases, $S_g$ selects group $g$, and $\epsilon=10^{-6}$ stabilizes the ratios. We hold $\gamma$ fixed during backpropagation.
\begin{equation}
\begin{split}
 \gamma&=\min\!\left(1,\frac{r_a}{\|\hat a\|_2+\epsilon},\,
      \min_g\frac{r_g}{\|S_gB\hat a\|_2+\epsilon}\right),\\
 a&=\gamma\hat a,\quad x^+=x+B_ra,\quad h^+=h+B_va.
\end{split}
\end{equation}
The depth and camera heads then read the updated reconstruction state. At the next AFC update, we reproject the candidates with these predictions and reselect supported 3D points using the original matches and covariances.

We apply the joint update after DiT blocks 10, 20, and 29. After block 30, the final readout samples video features at the projected candidates. The reconstruction proposal head maps those features to a residual update. We scale the update to satisfy all reconstruction group bounds and add it to $x$ before Gaussian decoding (Sec.~\ref{sec:video_prior}).

\par\smallskip\noindent
\begin{minipage}{\linewidth}
{\small
\hrule\smallskip
\noindent\textbf{Algorithm S1. One AFC update}\par
\noindent\textbf{Input} Current reconstruction and video states $x,h$, and pixel matches $(u_i,u_j)$ with covariances $\Sigma_j$ computed before the first AFC update.
\begin{enumerate}
\setlength{\itemsep}{0pt}
\setlength{\parskip}{0pt}
  \item Read depth and cameras from $x$. Project and score the candidate planes, sample their features, and retain supported 3D points.
  \item Aggregate the aligned features using $w_k$. Pass the current tokens and aligned features to the group MLPs. Form $M_r,M_v$ and normalize each pair of output directions.
  \item Form $W,T$, evaluate $e,q$, and batch the JVPs through the reconstruction readouts with observations fixed.
  \item Form $H,b$, solve for $\hat a$ by Cholesky factorization, and apply the shared scale $\gamma$. Return $x+B_ra$ and $h+B_va$ with $a=\gamma\hat a$.
\end{enumerate}
\smallskip\hrule
\par}
\end{minipage}\par\smallskip

\subsection{Appearance feature bridge}
\label{sec:supp_bridge}
The appearance feature bridge sits between the reconstruction backbone and the Gaussian decoder (Sec.~\ref{sec:obs_preserving}). It uses patch-to-pixel lifting inspired by GrapeSplat~\citep{lu2026grapesplat} to turn reconstruction tokens into a dense appearance map for each input view. The decoder uses these maps to preserve local image detail in the 3D Gaussians.

\section{Experimental details}
\label{sec:supp_experimental_settings}

\subsection{Stage I: reconstruction supervision}
\label{sec:supp_stage1}

Stage~I trains the reconstruction stream with RGB, depth, normal, and relative roof-height supervision (Eq.~\ref{3}).
Following AnySplat~\citep{jiang2025anysplat}, we use the same sampled views as reconstruction inputs and as targets for RGB rendering supervision.
We keep the image encoder and camera head fixed and train the depth decoder, appearance feature bridge, Gaussian decoder, and relative-height predictor.
We train the appearance feature bridge and Gaussian decoder jointly with the mean $\ell_1$ color error on the input views (Sec.~\ref{sec:supp_bridge}). MatrixCity provides most of the geometric supervision.

\paragraph{Training schedule.}
We train for 8,000 steps with four scenes per batch. Table~\ref{tab:supp_hyperparameters} summarizes the training settings.

\begin{figure}[htbp]
    \centering
    \includegraphics[width=.80\linewidth]{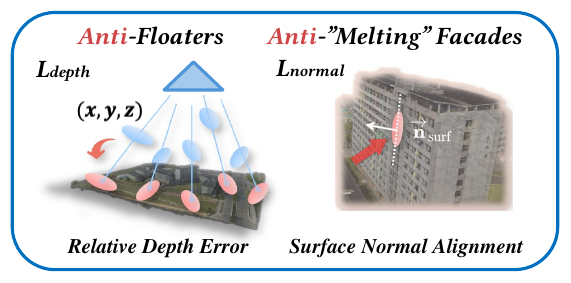}
    \caption{\textbf{Depth and normal supervision in Stage~I.} The depth loss constrains surface position along viewing rays. The normal loss aligns the shortest Gaussian axes with reference surface orientations.}
    \label{fig:stage1_geometry}
\end{figure}

\begin{samepage}
\paragraph{Depth: where is the surface?}
The left panel of Fig.~\ref{fig:stage1_geometry} illustrates how depth supervision guides Gaussians toward the observed surface.
We compare rendered depth $\hat D$ with reference depth $D$. We measure error relative to $D$, so the same absolute error carries more weight for a nearby surface.
\begin{equation}
 \mathcal L_{\mathrm{depth}}=\frac{1}{|\mathcal V_D|}
 \sum_{u\in\mathcal V_D}\frac{|\hat D(u)-D(u)|}{D(u)+\epsilon}.
\end{equation}
\end{samepage}

\paragraph{Normals: which way does the surface face?}
The right panel shows why surface orientation matters for facades. We use the shortest covariance axis of each Gaussian as its normal.
For rendered and reference normals $\hat n$ and $n$, the loss is
\begin{equation}
 \mathcal L_{\mathrm{normal}}=\frac{1}{|\mathcal V_N|}
 \sum_{u\in\mathcal V_N}\bigl(1-|\hat n(u)^\top n(u)|\bigr).
\end{equation}
The absolute dot product treats opposite signs of the same axis as aligned.
The masks $\mathcal V_D$ and $\mathcal V_N$ keep pixels with a valid label for that loss and sufficient rendered opacity.

\paragraph{Relative roof height.}
MatrixCity provides depth maps and calibrated cameras that locate roof points in 3D.
For valid roof patches, we use the reference roof point to measure the offset along the up direction from the same local anchor used for the prediction.
The height loss compares the two offsets in the same scene coordinates.

\subsection{Stage II: video adaptation}
\label{sec:supp_stage2}

We initialize the video backbone from \mbox{Wan2.2-TI2V-5B}. In Stage~II, we train the video LoRA adapters and AFC together while keeping the Stage~I reconstruction backbone fixed.
We adapt the video branch on aerial RGB views, primarily from GoogleEarth and UrbanScene3D (Sec.~\ref{sec:experimental_setup}).
At a fixed noise level, the video branch learns to predict the correction $z-z_t$ from the noisy image latent $z_t$ to the clean latent $z$.
The reconstruction state provides context for this prediction.
The two streams exchange features at three bidirectional coupling layers, followed by a final readout from the video stream to the reconstruction stream. Inference follows this fixed order.

\subsection{Stage II: rendering and auxiliary losses}
\label{sec:supp_refinement}

The model takes a sparse subset of views from each training sequence as input.
We supervise RGB renderings at the input views and, when available, at held-out target views from the same sequence.

The pair loss uses the final rendered depth to reproject each source match in both directions. It applies a robust penalty with an 8-pixel scale.
The coverage loss penalizes opacity below 0.5 at the same pixels.
The match weights account for repeated use of the same source pixel.

The auxiliary losses in Eq.~\ref{stage2} supervise video denoising, align the exchanged video features with clean source-image VAE latents, and penalize large updates.
Where geometry labels are available, we also supervise depth and normals and compare candidate intersections with the reference geometry.
These losses train model parameters shared across scenes.

\subsection{Inputs, cameras, and target views}
\label{sec:supp_evaluation}

\paragraph{Preprocessing.}
The reconstruction stream uses $448\times448$ input crops.
We crop and resize each RGB image, its masks, and its reference maps together, and update the camera intrinsics for the crop and resize.
We use bilinear interpolation for RGB and nearest-neighbor resampling for the depth validity mask.
We renormalize interpolated normals to unit length and keep invalid reference pixels masked.
Both streams use the same image region and view order. The video branch resizes the crop for the Wan VAE, which encodes it into a smaller latent map. The DiT groups the latent cells into an $8\times8$ token grid for each view. AFC resamples these tokens to its $16\times16$ video sampling grid.

\paragraph{Camera protocol for novel-view evaluation.}
For the pose-free comparison on GoogleEarth and UAVScenes, we follow AnySplat~\citep{jiang2025anysplat}. Each method reconstructs the scene from the input views. A second pass with the input and target images predicts the target cameras for that reconstruction. The input cameras shared by the two passes align the translation scale. Section~\ref{sec:supp_fixed_cameras} gives the protocol for comparing reconstructions at shared target cameras.

\paragraph{Input and target views.}
On GoogleEarth and UAVScenes, we use the LLFF split with \texttt{llffhold=2}. We hold out every other frame for evaluation and select 2, 4, or 16 input images from the remaining frames.
All methods use the same input and target images for each sample.
On MatrixCity, the comparison with per-scene optimization uses 32 or 64 input views.
The optimization methods use the poses and initialization required by their published pipelines, and we count both initialization and fitting in their reconstruction times.
For the feed-forward comparison, we use direct predictions with per-scene refinement disabled, including TokenBA for ZipSplat.

\paragraph{Baseline fine-tuning.}
We fine-tune the feed-forward baselines on the same curated aerial training split as AnyCity.
We follow each baseline's training recipe, using the published description to reproduce the setup when a ready-to-use training configuration is unavailable.

\paragraph{Paired flight settings.}
AnyCity-Aerial compares two altitudes (150 and 250~m) and, in a separate experiment, two viewing angles ($0^\circ$ nadir and $45^\circ$ oblique).
Each setting uses four input views. We align the predicted input camera poses to the benchmark poses and render both settings at the same benchmark target cameras (Sec.~\ref{sec:anycity_aerial_eval}).
To measure the effect of the geometric terms during inference, we set $\lambda=\mu=0$ and keep the trained AnyCity weights fixed.
We report this comparison separately from ablations that retrain the model.

\subsection{Metrics and observation regions}
\label{sec:supp_metrics}

\paragraph{RGB metrics.}
We compare rendered and reference images using PSNR, SSIM, and LPIPS.
Table~\ref{tab:quantitative} reports these scores on target views, with reconstruction time also reported for MatrixCity.
Table~\ref{tab:anycity_aerial_eval} reports full-image PSNR for the flight comparisons.
Table~\ref{tab:ablation}(a) reports PSNR in both observation regions and LPIPS in the weakly observed region.

\paragraph{Depth metrics.}
The GoogleEarth ablations use MoGe2-Aerial depth as the reference. The AnyCity-Aerial flight comparisons use depth from the reference reconstruction that supplies the controlled renders.
Let $\mathcal V$ contain the pixels with valid reference depth $D$ and $\mathcal V_+\subseteq\mathcal V$ those with finite, positive predicted depth $\hat D$.
AbsRel averages the relative depth error over $\mathcal V_+$. For $\delta_1$, a reference-valid pixel counts as correct if its predicted depth is valid and within a factor of 1.25 of the reference:
\begin{equation}
\begin{split}
 \operatorname{AbsRel}&=\frac1{|\mathcal V_+|}\sum_{u\in\mathcal V_+}
 \frac{|\hat D(u)-D(u)|}{D(u)},\\
 \delta_1&=\frac{100}{|\mathcal V|}\sum_{u\in\mathcal V_+}
 \mathbf1\!\left[\max\!\left(\frac{\hat D(u)}{D(u)},\frac{D(u)}{\hat D(u)}\right)<1.25\right].
\end{split}
\end{equation}
Table~\ref{tab:ablation}(b) reports both depth metrics. Table~\ref{tab:anycity_aerial_eval} reports AbsRel separately for the two observation regions.

\paragraph{Observation regions.}
We call a valid reference point observation-supported if at least two input cameras see it with sufficient parallax.
In a paired flight comparison, this condition must hold in both settings.
The remaining valid points form the weakly observed region.
All methods use the same masks for a given comparison.

\subsection{Training and inference settings}
\label{sec:supp_control_settings}

\paragraph{Training setup.}
We use AdamW in both stages. Table~\ref{tab:supp_hyperparameters} gives the settings for training and the Stage~II controls.

\begin{table}[htbp]
  \centering
  \caption{\textbf{Training hyperparameters and compute.}}
  \label{tab:supp_hyperparameters}
  \small
  \setlength{\tabcolsep}{3pt}
  \begin{tabular}{@{}>{\raggedright\arraybackslash}p{0.50\linewidth}
                    >{\raggedright\arraybackslash}p{0.45\linewidth}@{}}
    \toprule
    \textbf{Setting} & \textbf{Value} \\
    \midrule
    \multicolumn{2}{@{}l}{\textit{Stage I}} \\
    Steps & 8,000 \\
    Batch size & 4 scenes \\
    Input views per scene & 2--24, randomly sampled~\citep{jiang2025anysplat} \\
    Peak learning rate & $10^{-4}$ \\
    Learning-rate schedule & 500-step linear warmup, then cosine decay \\
    Loss weights $(\lambda_D,\lambda_N,\lambda_H)$ & $(0.1,0.1,0.1)$ \\
    \addlinespace[2pt]
    \multicolumn{2}{@{}l}{\textit{Stage II}} \\
    Steps & 10,000 \\
    AFC learning rate & $10^{-4}$ \\
    Video adapter learning rate & $3\times10^{-5}$ \\
    LoRA rank $r$ and alpha $\alpha$ & 16 and 16 \\
    \addlinespace[2pt]
    \multicolumn{2}{@{}l}{\textit{Stage II controls}} \\
    Weight decay & 0.01 \\
    Gradient norm clip & 1 \\
    \addlinespace[2pt]
    \multicolumn{2}{@{}l}{\textit{Compute}} \\
    Feature computation & bf16 \\
    Geometric accumulation and solve & FP32 \\
    Training hardware & 8 A800 GPUs (80~GB each) \\
    \bottomrule
  \end{tabular}
\end{table}

\paragraph{Matched training runs.}
The training controls use the Stage~II protocol for data and optimization. Each control changes the component or trainable parameters specified in Sec.~\ref{sec:supp_attribution}.

\paragraph{Inference cost.}
The full reconstruction pass encodes the input views, matches them once, and runs three AFC updates with JVPs and coefficient solves. The final video readout and Gaussian decoder then produce the scene.

With $N$ input views, matching considers $N(N-1)$ ordered pairs, while the number of AFC updates and the number of JVPs per update stay fixed.
On MatrixCity, Table~\ref{tab:quantitative}(b) reports reconstruction times of 3.3~s with 32 input views and 4.4~s with 64. Inference uses 22.34~GB of GPU memory on one NVIDIA A800.

\section{Discussion}
\label{sec:supp_questions}

\paragraph{Why sample more than one surface?}
Different roof heights can project to almost the same image patch. Features sampled at those heights can all describe the same roof. Combining them gives AFC useful appearance cues even when the roof's exact depth remains uncertain.

\paragraph{How does AFC update both streams?}
Image matches constrain where a surface should project, and the selected 3D points constrain where it should lie in space. These cues guide one set of update coefficients for both the reconstruction and video streams (Eq.~\ref{eq:anycity_objective}). Patches with uncertain depth can still supply features to the update proposals.

\paragraph{What evidence supports AFC's reference points?}
The initial depth and cameras place the search window. Image features select the match within it. AFC reprojects the candidate planes at each update and checks them against that same match, keeping the image target fixed as the geometry changes. AFC uses a 3D point when the views can distinguish the surface from nearby alternatives and give the leading candidate group enough support.

At 30\% coverage on MatrixCity, full selection achieves 84.5\% accuracy against the dataset's reference depth, compared with 71.4\% for reprojection alone (Table~\ref{tab:supp_reliability}). The perturbation study also measures 3D error against MatrixCity depth. AFC reduces that error across the depth sweep and at the two smaller pose perturbations (Sec.~\ref{sec:supp_initial_perturbation}).

\paragraph{What does Stage II add?}
Stage~II adapts the video features to aerial imagery and trains AFC to use those features in the reconstruction. Compared with Stage~I, the full model improves PSNR on GoogleEarth by 2.14~dB in observation-supported regions and 2.43~dB in weakly observed regions (Table~\ref{tab:ablation}(a)). The LoRA ablation isolates the contribution of adapting the video branch to aerial imagery (Table~\ref{tab:ablation}(b)).

\paragraph{When does AFC run?}
AFC runs three times in a single forward pass. At each exchange, it solves a linear system and applies the shared coefficients to both streams. After the final DiT block, the reconstruction stream reads the video features once more before the decoder builds the Gaussian scene.

\begin{figure}[htbp]
    \centering
    \includegraphics[width=\linewidth]{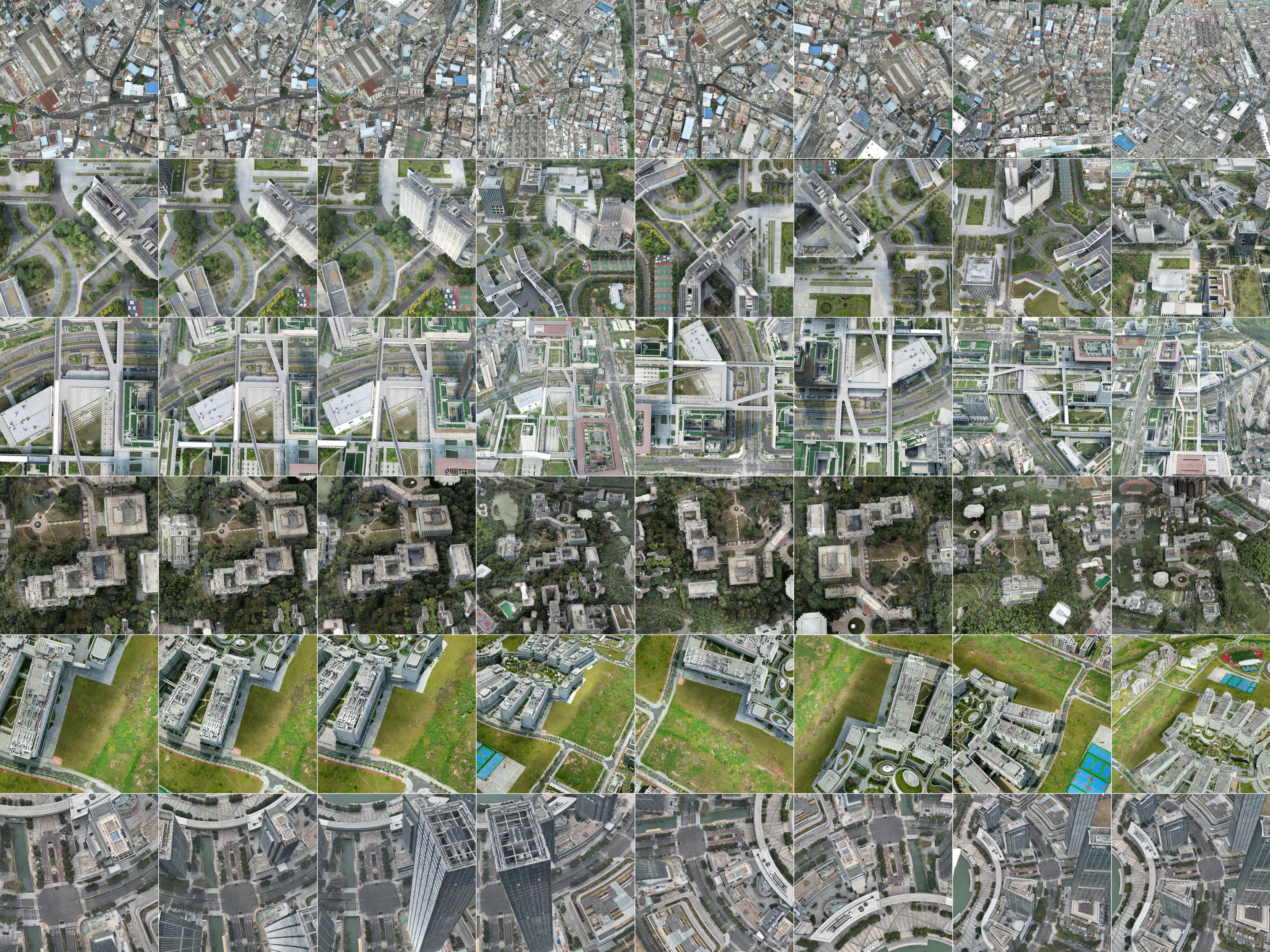}
    \caption{\textbf{AnyCity-Aerial benchmark views.} Each row shows eight aerial views of one of the six scenes.}
    \label{fig:supp_dataset}
\end{figure}

\paragraph{Why do more views affect the metrics differently?}
From four to sixteen input views, PSNR rises on both GoogleEarth and UAVScenes, while SSIM falls and LPIPS rises (Table~\ref{tab:quantitative}(a)). More views can cover parts of the scene that were previously unseen and reduce large color errors. Views along the same flight path often show a building from similar angles, so even small alignment errors can blur roof edges and window patterns when their features are combined. This can explain why the overall pixel error falls even as fine details become less sharp.

\section{Dataset presentation}
\label{sec:supp_dataset}

\subsection{Training and evaluation splits}
Table~\ref{tab:supp_data_splits} shows how each dataset is used. We use disjoint UrbanScene3D scenes for Stage~II training and AnyCity-Aerial.

\begin{table}[htbp]
  \centering
  \caption{\textbf{Training and evaluation datasets.}}
  \label{tab:supp_data_splits}
  \small
  \setlength{\tabcolsep}{3pt}
  \begin{tabular}{@{}p{0.24\linewidth}>{\raggedright\arraybackslash}p{0.72\linewidth}@{}}
    \toprule
    \textbf{Dataset} & \textbf{Training and evaluation use} \\
    \midrule
    MatrixCity & Main source of Stage~I geometric supervision. Evaluation uses held-out aerial data. \\
    GoogleEarth & CityDreamer data for Stage~II and a held-out split for novel-view evaluation. \\
    UrbanScene3D & Real UAV scenes for Stage~II, with disjoint scenes used to build AnyCity-Aerial. \\
    UAVScenes & Novel-view evaluation. Every other frame is held out. We select 2, 4, or 16 inputs from the remaining frames. \\
    AnyCity-Aerial & Six benchmark scenes from DJI captures, GauU-Scene, and UrbanScene3D for paired flight evaluation. \\
    \bottomrule
  \end{tabular}
\end{table}

\subsection{AnyCity-Aerial configuration}
AnyCity-Aerial pairs input flights that differ in altitude or camera tilt. Each pair shares the same scene and target cameras. Table~\ref{tab:supp_dataset} summarizes the benchmark, and Fig.~\ref{fig:supp_dataset} shows example views.

\begin{table}[htbp]
    \centering
    \caption{\textbf{AnyCity-Aerial benchmark.}}
    \label{tab:supp_dataset}
    \small
    \setlength{\tabcolsep}{4pt}
    \begin{tabular}{@{}p{0.22\linewidth}p{0.71\linewidth}@{}}
        \toprule
        \textbf{Item} & \textbf{Description} \\
        \midrule
        Data sources & DJI UAV captures and scenes from GauU-Scene and UrbanScene3D. \\
        Scale & 6 scenes, 288 sequences, and 18,432 frames (64 frames per sequence on average). \\
        Flight altitude & Five levels per scene. Table~\ref{tab:anycity_aerial_eval} compares 150 and 250~m. \\
        Camera tilt & Five levels per scene. Table~\ref{tab:anycity_aerial_eval} compares $0^\circ$ (nadir) and $45^\circ$ (oblique). \\
        Flight paths & We use recorded altitude, gimbal pitch, and GPS tracks to design parallel and orbiting camera paths. \\
        Sequence generation & We reconstruct each scene as 3D Gaussians and render aerial sequences along the controlled camera paths. \\
        \bottomrule
    \end{tabular}
\end{table}

\section{Additional experiments}
\label{sec:supp_extra}

To understand how AFC uses sparse aerial observations to guide reconstruction, we conduct further experiments. We ablate the geometric constraints on MatrixCity (Sec.~\ref{sec:supp_matrix_geometry}), examine the sources of the Stage~II gain (Sec.~\ref{sec:supp_attribution}), and test the reliability of the selected 3D reference points (Sec.~\ref{sec:supp_reliability}). We also compare AnyCity's reconstructed geometry with Gen3R's (Sec.~\ref{sec:supp_generative_comparison}) and test the final readout and shared update coefficients (Sec.~\ref{sec:supp_final_read}). Sections~\ref{sec:supp_fixed_cameras} and~\ref{sec:supp_initial_perturbation} cover evaluation with shared target cameras and under perturbations to the initial geometry, respectively.

\begingroup
\newenvironment{suppcomparison}[2]{%
  \par\medskip\noindent\begin{minipage}{\linewidth}
  \centering
  \captionsetup{hypcap=false}
  \captionof{table}{#1}\label{#2}
  \small
  \setlength{\tabcolsep}{4pt}\renewcommand{\arraystretch}{1.10}%
}{\end{minipage}\par\medskip}

\subsection{Geometry ablations on MatrixCity}
\label{sec:supp_matrix_geometry}

\begin{figure}[htbp]
  \centering
  \newlength{\matrixpanelwidth}
  \setlength{\matrixpanelwidth}{\dimexpr(.88\linewidth-8pt)/2\relax}
  \begin{minipage}[c]{.12\linewidth}
    \footnotesize
    \begin{tabular}{@{}r@{}}
      \parbox[c][.25\matrixpanelwidth][c]{\linewidth}{\raggedleft RGB} \\
      \parbox[c][.25\matrixpanelwidth][c]{\linewidth}{\raggedleft Depth head} \\
      \parbox[c][.25\matrixpanelwidth][c]{\linewidth}{\raggedleft Gaussian depth} \\
      \parbox[c][.25\matrixpanelwidth][c]{\linewidth}{\raggedleft GT depth}
    \end{tabular}
  \end{minipage}\hspace{4pt}%
  \begin{minipage}[c]{\matrixpanelwidth}
    \includegraphics[width=\linewidth]{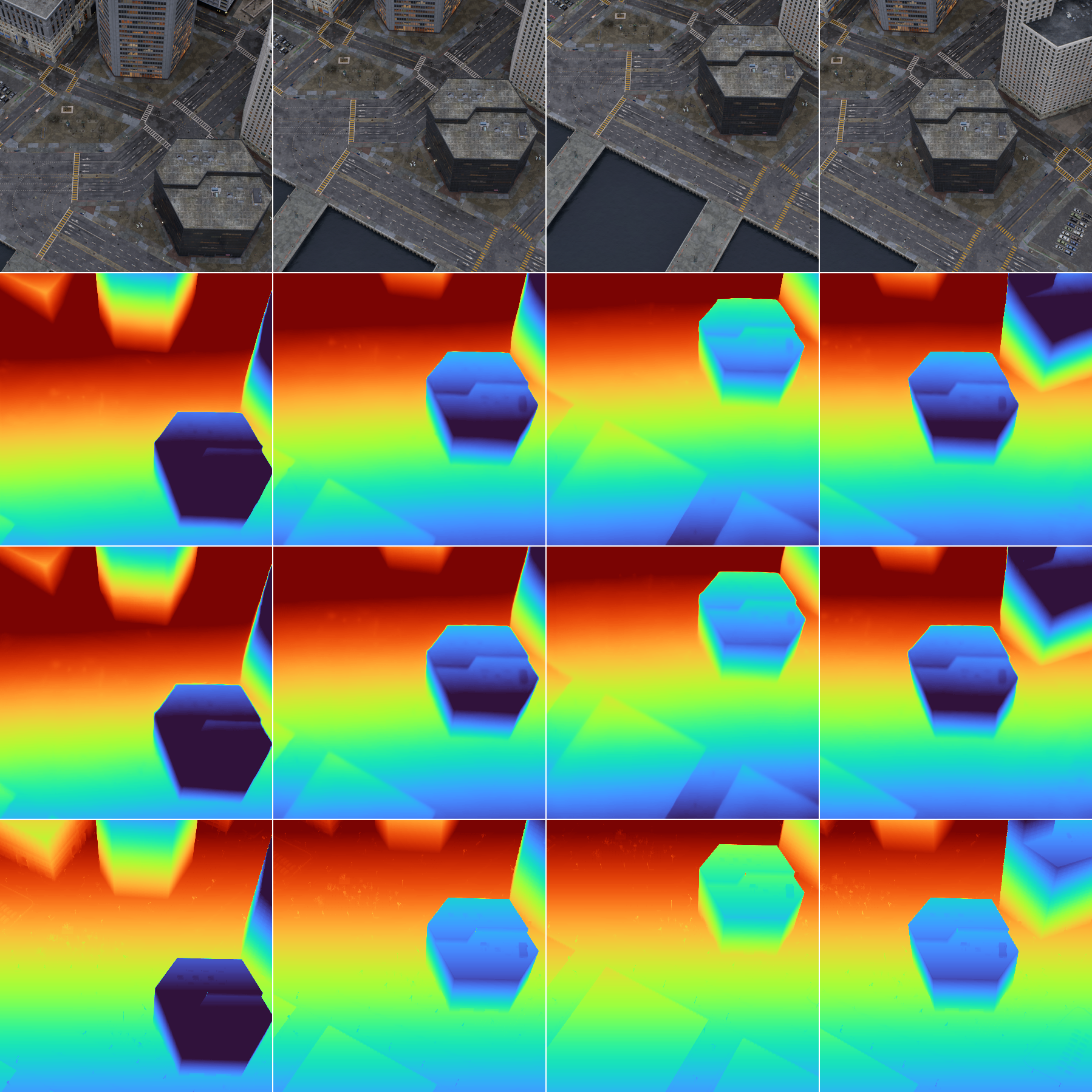}
  \end{minipage}\hspace{4pt}%
  \begin{minipage}[c]{\matrixpanelwidth}
    \includegraphics[width=\linewidth]{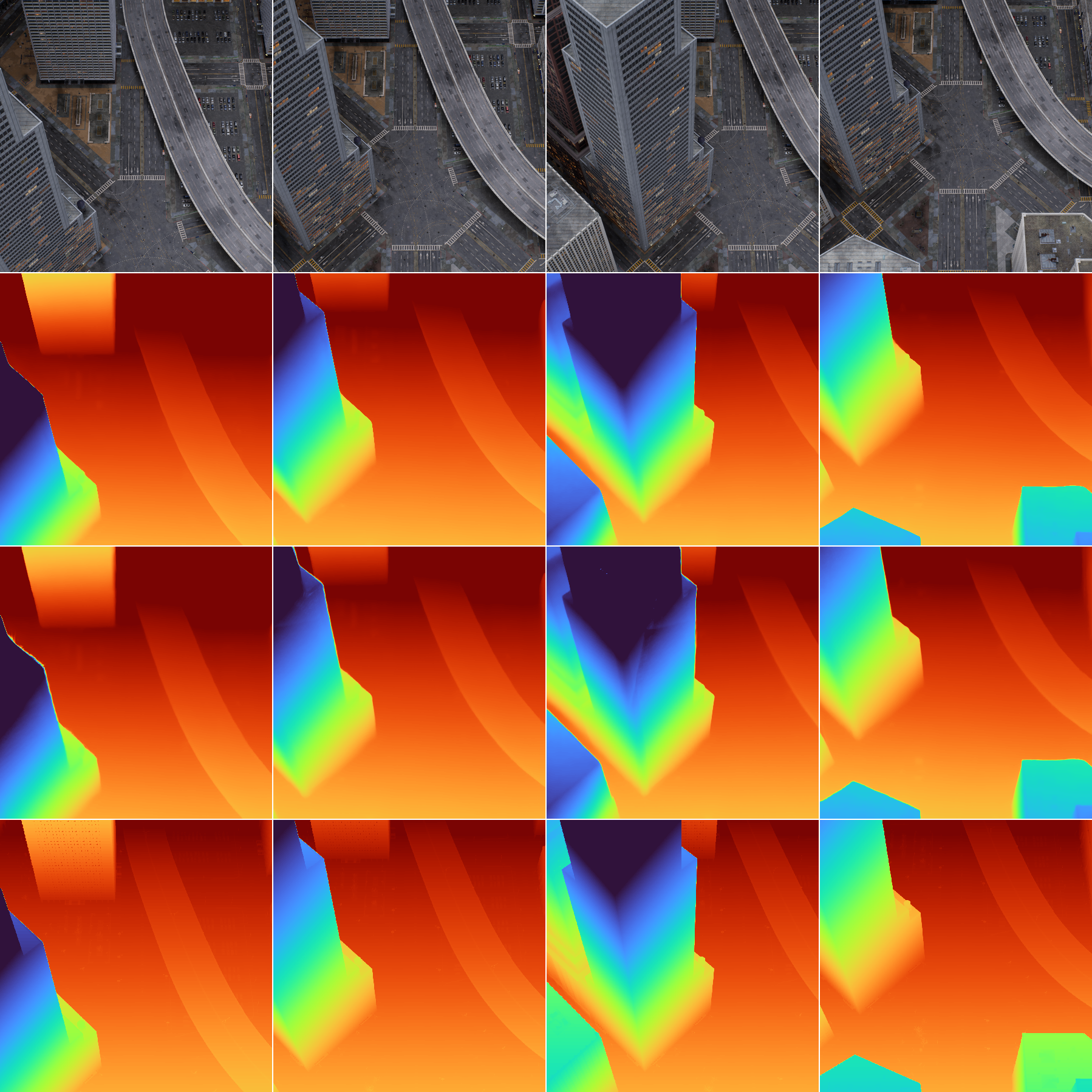}
  \end{minipage}
  \caption{\textbf{Depth visualization on MatrixCity.} Each column shows one view. From top to bottom, the rows show RGB, depth predicted by the depth head, depth rendered from the 3D Gaussians, and MatrixCity ground-truth depth.}
  \label{fig:supp_matrix_geometry}
\end{figure}

Table~\ref{tab:ablation}(b) evaluates geometry on GoogleEarth using depth predicted by MoGe2-Aerial. We repeat the two geometric-term ablations on MatrixCity's official aerial test split~\citep{li2023matrixcity}, using the dataset's rendered reference depth. We disable one term at a time at inference, keeping the checkpoint, four input views, and target cameras fixed.

\begin{suppcomparison}{\textbf{Four-view AFC ablations on MatrixCity.}}{tab:supp_matrix_geometry}
\begin{tabular*}{\linewidth}{@{\extracolsep{\fill}}lccccc@{}}
\toprule
Model configuration & PSNR$\uparrow$ & SSIM$\uparrow$ & LPIPS$\downarrow$ & \shortstack{$\delta_1\uparrow$\\(\%)} & AbsRel$\downarrow$ \\
\midrule
w/o reprojection ($\lambda=0$) & 21.03 & 0.597 & 0.457 & 91.6 & 0.126 \\
w/o 3D point term ($\mu=0$) & 20.96 & 0.593 & 0.463 & 90.9 & 0.129 \\
\midrule
\textbf{Full AnyCity} & \textbf{21.80} & \textbf{0.615} & \textbf{0.435} & \textbf{94.8} & \textbf{0.105} \\
\bottomrule
\end{tabular*}
\end{suppcomparison}

Removing either geometric term reduces depth accuracy and RGB rendering quality (Table~\ref{tab:supp_matrix_geometry}). Reprojection keeps the update consistent with observed pixel matches, whose depth constraints weaken with low parallax. Retained 3D reference points directly constrain surface positions where the views can distinguish them. AFC uses both signals to choose the shared update coefficients, so video features can refine the scene under the same geometric constraints as the reconstruction stream. Figure~\ref{fig:supp_matrix_geometry} compares depth from the prediction head and the Gaussians with the reference maps.

\subsection{Attributing the Stage II gain}
\label{sec:supp_attribution}

Table~\ref{tab:ablation}(a) measures the overall gain from Stage~II. The controls test additional aerial training, video pretraining, and the AFC interaction. Each Stage~II run starts from the same Stage~I reconstruction checkpoint and uses the same sample order and update budget.

\begin{suppcomparison}{\textbf{Stage II controls on GoogleEarth.} The Stage~I and Full AnyCity rows come from Table~\ref{tab:ablation}(a).}{tab:supp_training_controls}
\begin{tabular*}{\linewidth}{@{\extracolsep{\fill}}lccc@{}}
\toprule
Model configuration & \shortstack{Obs.\\PSNR$\uparrow$} & \shortstack{Weak\\PSNR$\uparrow$} & \shortstack{Weak\\LPIPS$\downarrow$} \\
\midrule
Stage~I (recon. only) & 21.48 & 17.83 & 0.304 \\
\addlinespace
Recon. adapters & 21.85 & 18.20 & 0.296 \\
Recon. fine-tuning & 22.21 & 18.53 & 0.289 \\
Video + cross-attention & 21.63 & 17.49 & 0.318 \\
AFC with random video & 22.05 & 18.39 & 0.295 \\
\midrule
\textbf{Full AnyCity} & \textbf{23.62} & \textbf{20.26} & \textbf{0.266} \\
\bottomrule
\end{tabular*}
\end{suppcomparison}

Standard cross-attention performs worse than Stage~I in weakly observed regions (Table~\ref{tab:supp_training_controls}). With weak parallax, similar-looking patches can come from different surfaces. Attention can transfer video detail to the wrong surface and distort the initial reconstruction. AFC samples at projected roof and facade candidates and weights the samples by observed matches and how well the views distinguish the candidates. Reprojection constrains the shared update, with 3D reference points added only where the views support their positions. The other controls also indicate gains from aerial training and video pretraining.

\subsection{Reference-point reliability}
\label{sec:supp_reliability}

We assess the 3D reference points from Sec.~\ref{sec:supp_surface_interaction} before the geometric update. For each accepted location, we back-project the MatrixCity reference depth and measure the distance to the selected point. We divide these distances by scene scale. Coverage is the fraction of eligible input locations each selection rule accepts.

\begin{suppcomparison}{\textbf{Reference-point selection on MatrixCity.} Accuracy is the percentage of accepted points within 2\% of scene scale from their reference positions.}{tab:supp_reliability}
\begin{tabular*}{\linewidth}{@{\extracolsep{\fill}}lccc@{}}
\toprule
Selection rule & \shortstack{Coverage\\(\%)} & \shortstack{Median 3D\\error$\downarrow$} & \shortstack{Accuracy\\(\%)$\uparrow$} \\
\midrule
Reprojection only & 30.0 & 0.0138 & 71.4 \\
Mode confidence only & 30.0 & 0.0164 & 64.3 \\
Shuffled confidence & 30.0 & 0.0162 & 59.0 \\
\textbf{Full selection} & \textbf{30.0} & \textbf{0.0090} & \textbf{84.5} \\
\bottomrule
\end{tabular*}
\end{suppcomparison}

At matched coverage, full selection lowers 3D error and improves accuracy (Table~\ref{tab:supp_reliability}). Similar reprojections can come from different surfaces, and a high mode score can occur with weak geometric support. AFC combines mode confidence with geometric support to retain positions the views can distinguish.

\subsection{Qualitative comparison with Gen3R}
\label{sec:supp_generative_comparison}

With the same aerial inputs and matched viewpoints (Fig.~\ref{fig:supp_generative}), AnyCity preserves more regular facade rows and clearer building and street boundaries. These examples illustrate the structures that AFC aims to preserve by grounding video-guided updates in cross-view observations.

\begin{figure}[htbp]
  \centering
  {\small\bfseries
    \makebox[0.39\linewidth]{Gen3R}%
    \makebox[0.39\linewidth]{AnyCity}\par}
  \smallskip
  \includegraphics[width=0.78\linewidth]{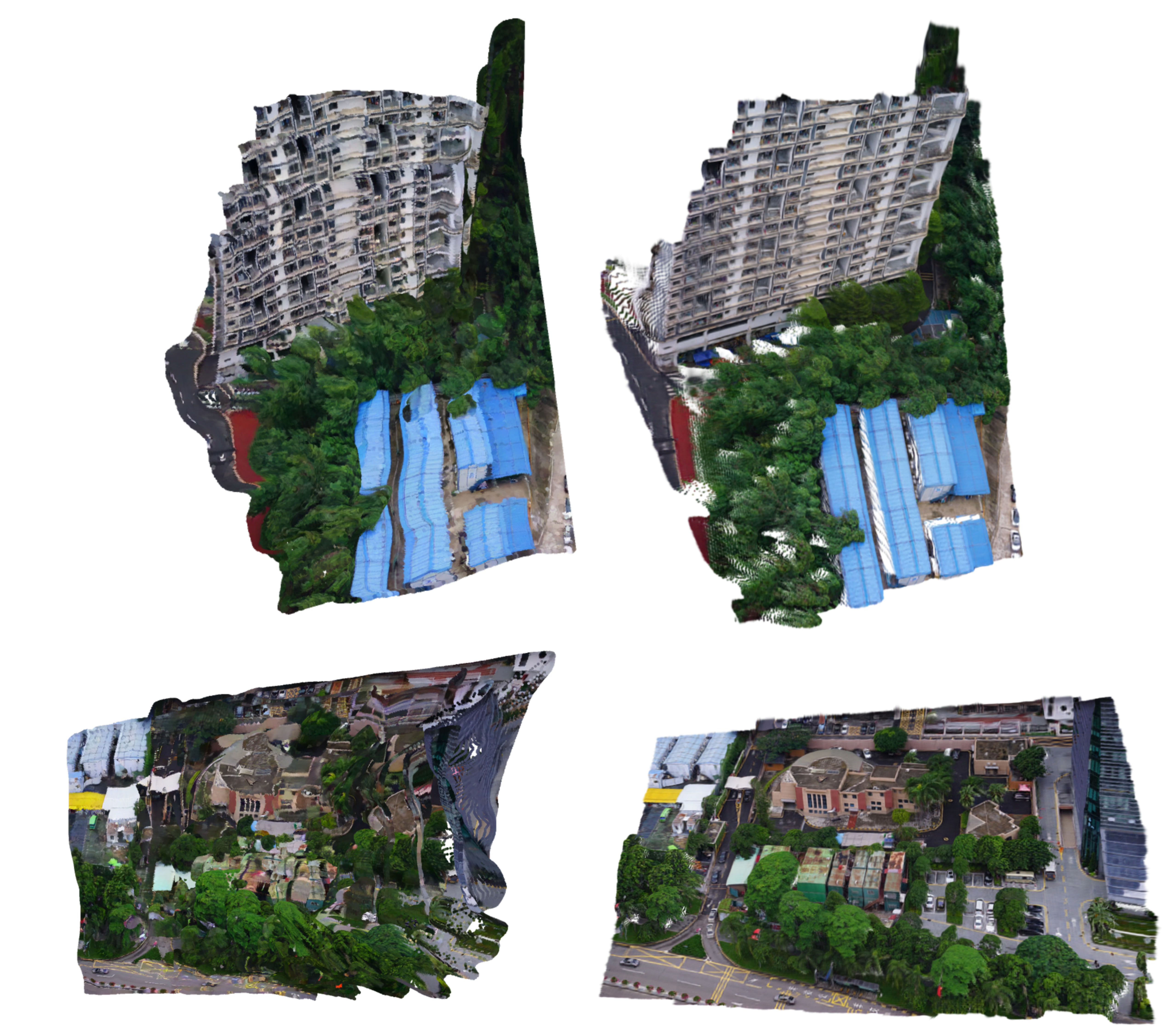}\par
  \caption{\textbf{3D reconstruction compared with Gen3R.}
    Each row shows the same scene reconstructed by Gen3R (left) and AnyCity (right).}
  \label{fig:supp_generative}
\end{figure}

\subsection{The final readout and coupling choices}
\label{sec:supp_final_read}

We test the final video-to-reconstruction readout after DiT block~30 with the same checkpoint and prediction heads. For the coefficient ablation, we retrain with separate coefficients for the two streams and use the same direction-generating architecture and training schedule.

\begin{suppcomparison}{\textbf{Final readout and coupling ablations.}}{tab:supp_final_read}
\begin{tabular*}{\linewidth}{@{\extracolsep{\fill}}lccc@{}}
\toprule
Model configuration & \shortstack{Reproj. error$\downarrow$\\(px)} & \shortstack{$\delta_1\uparrow$\\(\%)} & AbsRel$\downarrow$ \\
\midrule
\multicolumn{4}{@{}l}{\textit{Same-checkpoint inference}} \\
\textbf{Full AnyCity} & 1.84 & \textbf{93.8} & \textbf{0.120} \\
No final readout & 2.06 & 92.9 & 0.128 \\
Geometry-constrained readout & \textbf{1.73} & 93.5 & 0.123 \\
\addlinespace
\multicolumn{4}{@{}l}{\textit{Matched Stage~II retraining}} \\
Separate coefficients & 2.17 & 92.6 & 0.132 \\
\bottomrule
\end{tabular*}
\end{suppcomparison}

Removing the final readout worsens reprojection and target depth (Table~\ref{tab:supp_final_read}), suggesting that the final video features add useful information after the intermediate coupling steps. Applying geometric constraints at the readout improves reprojection further at a small cost to target depth accuracy. This trade-off supports using AFC to constrain the intermediate updates and the final readout to incorporate the remaining video cues. Separate coefficients worsen reprojection and depth, supporting the use of shared coefficients to guide both streams with the same geometric evidence.

\subsection{Fixed-camera novel-view synthesis}
\label{sec:supp_fixed_cameras}

On held-out GoogleEarth scenes, AnyCity and the fine-tuned SplatWeaver and AnySplat baselines reconstruct from the same input RGB images. We use nested sets of 2, 4, and 16 inputs and keep the target frames fixed. For each reconstruction, the same input-only similarity alignment uses the input camera orientations to fit rotation and their centers to fit scale and translation. All methods render with the same benchmark target poses and intrinsics.

\begin{suppcomparison}{\textbf{Fixed-camera comparison on GoogleEarth.} Each view-count column reports PSNR$\uparrow$ / SSIM$\uparrow$ / LPIPS$\downarrow$. $\dagger$ denotes fine-tuning on the aerial training split.}{tab:supp_fixed_cameras}
\begin{tabular*}{\linewidth}{@{\extracolsep{\fill}}lccc@{}}
\toprule
Method & 2 views & 4 views & 16 views \\
\midrule
SplatWeaver$^\dagger$ & 15.4/0.31/0.42 & 16.6/0.35/0.38 & 15.0/0.22/0.52 \\
AnySplat$^\dagger$ & 15.0/0.28/0.37 & 16.3/0.32/0.34 & 14.7/0.21/0.51 \\
\textbf{AnyCity} & \textbf{17.6/0.33/0.29} & \textbf{20.7/0.46/0.21} & \textbf{21.1/0.45/0.26} \\
\bottomrule
\end{tabular*}
\end{suppcomparison}

\subsection{Initial geometry perturbation}
\label{sec:supp_initial_perturbation}

To test how errors in the initial geometry affect AFC's ability to refine the reconstruction, we perturb the predicted depth and camera poses in separate sweeps. We use four input views per scene on the MatrixCity test split. The depth sweep adds a Gaussian offset to the log depth of each view. The pose sweep rotates each camera about a random axis and shifts its center in a random direction. We use the median valid reference depth as the scene scale, and we compute pixel matches from the perturbed initial geometry and keep the sampled offsets fixed throughout the AFC updates.

We evaluate the geometry before the first AFC update and after the third, before the final readout. Both snapshots use the same alignment, fitted from the unperturbed input cameras. We measure the median 3D distance to points back-projected from MatrixCity depth at the same pixels and divide it by scene scale. We measure reference-point accuracy as in Sec.~\ref{sec:supp_reliability} and report the coverage returned by the selection rule on the same eligible input locations at both snapshots.

\begin{suppcomparison}{\textbf{Initial geometry perturbation on MatrixCity.} Depth levels give the log-depth noise standard deviation. Pose levels give the rotation magnitude and camera-center shift as a percentage of scene scale.}{tab:supp_initial_perturbation}
\begin{tabular*}{\linewidth}{@{\extracolsep{\fill}}lcccccc@{}}
\toprule
& \multicolumn{2}{c}{3D error$\downarrow$} & \multicolumn{2}{c}{Acc. (\%)$\uparrow$} & \multicolumn{2}{c}{Coverage (\%)} \\
\cmidrule(lr){2-3}\cmidrule(lr){4-5}\cmidrule(lr){6-7}
Perturbation & Before & After & Before & After & Before & After \\
\midrule
None & 0.010 & 0.008 & 84.24 & 86.70 & 30.20 & 31.93 \\
Depth $0.025$ & 0.013 & 0.011 & 74.21 & 79.10 & 26.88 & 28.86 \\
Depth $0.05$ & 0.018 & 0.015 & 58.15 & 64.90 & 21.12 & 23.02 \\
Depth $0.10$ & 0.034 & 0.032 & 35.11 & 38.10 & 11.89 & 11.28 \\
Pose $1^\circ$, $0.5\%$ & 0.016 & 0.014 & 63.06 & 69.04 & 24.85 & 26.10 \\
Pose $2^\circ$, $1\%$ & 0.028 & 0.024 & 44.95 & 48.48 & 17.16 & 16.11 \\
Pose $4^\circ$, $2\%$ & 0.055 & 0.056 & 23.87 & 22.79 & 8.08 & 7.18 \\
\bottomrule
\end{tabular*}
\end{suppcomparison}

AFC reduces 3D error and improves reference-point accuracy across the depth sweep and at the two smaller pose perturbations (Table~\ref{tab:supp_initial_perturbation}). At the largest pose perturbation, 3D error rises slightly and both accuracy and coverage decline. Large pose shifts can move true correspondences outside the initial search window, limiting the evidence available to the update.

\endgroup

\section{Limitations and future work}
\label{sec:supp_discussion}

\subsection{Limitations}
When the input cameras are close together, roofs at different heights can appear in nearly the same place in each image. The images then provide little evidence about the true roof height, so AFC keeps fewer 3D reference points to guide the update. The video branch can still sharpen visual details, but the input views give little information about where those details belong in 3D.

\subsection{Future work}
A first direction is to use the reconstruction to plan the next flight pass. An initial scene can show which surfaces the cameras have seen. A planner could then choose viewpoints that reveal other surfaces and capture the same building from more varied angles.

Another direction is to extend the surface candidates in AFC. Curved patches and candidates at finer scales could better represent features such as curved roofs and thin structures. Gaussians that change over time could represent objects that move between flights.

\subsection{Downstream uses}
Survey teams can view the reconstructed scene from new angles and mark areas of interest. The same scene can serve as a visual record for urban surveys and documentation.

For tasks such as infrastructure inspection and change detection, the reconstruction can be paired with calibrated measurements to check dimensions and changes at the required level of accuracy.

\subsection{Data use and privacy}
Aerial images can contain details about private locations. We follow the permissions for each source when collecting or sharing images. Third-party images, models, and code retain their original licenses.

\section{Relation to Prior Work}
\label{sec:supp_component_lineage}

Gen3R~\citep{huang2026gen3r} uses video diffusion to generate aligned appearance and geometry latents, then decodes them into RGB frames and 3D scene attributes. AnyCity starts from a feed-forward reconstruction. AFC aligns features across views with projected roof and facade candidates. Pixel matches and supported 3D points constrain how AFC updates the two streams, and the Gaussian decoder turns the updated reconstruction into a renderable scene.

We follow AnySplat~\citep{jiang2025anysplat} in rendering RGB and depth from decoded Gaussians. Our design draws inspiration from GrapeSplat~\citep{lu2026grapesplat}, which aggregates multi-view features in a shared 3D voxel grid. Its patch-to-pixel lifting also inspires the appearance feature bridge that supplies dense per-view features to our Gaussian decoder.